\documentclass[letterpaper]{article} % DO NOT CHANGE THIS
\usepackage{aaai25}  % DO NOT CHANGE THIS
\usepackage{times}  % DO NOT CHANGE THIS
\usepackage{helvet}  % DO NOT CHANGE THIS
\usepackage{courier}  % DO NOT CHANGE THIS
\usepackage[hyphens]{url}  % DO NOT CHANGE THIS
\usepackage{graphicx} % DO NOT CHANGE THIS
\usepackage{natbib}  % DO NOT CHANGE THIS AND DO NOT ADD ANY OPTIONS TO IT
\usepackage{caption} % DO NOT CHANGE THIS AND DO NOT ADD ANY OPTIONS TO IT
\usepackage{algorithm}
\usepackage{algorithmic}

\usepackage{amsmath} % \begin{align}
\usepackage{amssymb} % \mathbb
\usepackage{amsfonts}
\usepackage{bm} % \bm
\usepackage{subfigure} % \subfigure
\usepackage{multirow} % \multirow
\usepackage{booktabs}
\usepackage{graphicx}
\usepackage{utfsym}
\usepackage{makecell}
\usepackage{newfloat}
\usepackage{listings}
\DeclareCaptionStyle{ruled}{labelfont=normalfont,labelsep=colon,strut=off} % DO NOT CHANGE THIS
\floatstyle{ruled}
\newfloat{listing}{tb}{lst}{}
\floatname{listing}{Listing}
\title{Data-Free Black-Box Federated Learning via Zeroth-Order Gradient Estimation}
\author{
    Xinge Ma, Jin Wang\thanks{Corresponding author.}, Xuejie Zhang\\
}
\affiliations{
    School of Information Science and Engineering\\
    Yunnan University\\
    Kunming, China\\
    maxinge@mail.ynu.edu.cn, \{wangjin, xjzhang\}@ynu.edu.cn
}

\usepackage{bibentry}
\begin{document}

\maketitle

\begin{abstract}
Federated learning (FL) enables decentralized clients to collaboratively train a global model under the orchestration of a central server without exposing their individual data. However, the iterative exchange of model parameters between the server and clients imposes heavy communication burdens, risks potential privacy leakage, and even precludes collaboration among heterogeneous clients. Distillation-based FL tackles these challenges by exchanging low-dimensional model outputs rather than model parameters, yet it highly relies on a task-relevant auxiliary dataset that is often not available in practice. Data-free FL attempts to overcome this limitation by training a server-side generator to directly synthesize task-specific data samples for knowledge transfer. However, the update rule of the generator requires clients to share on-device models for white-box access, which greatly compromises the advantages of distillation-based FL. This motivates us to explore a data-free and black-box FL framework via Zeroth-order Gradient Estimation (FedZGE), which estimates the gradients after flowing through on-device models in a black-box optimization manner to complete the training of the generator in terms of fidelity, transferability, diversity, and equilibrium, without involving any auxiliary data or sharing any model parameters, thus combining the advantages of both distillation-based FL and data-free FL. Experiments on large-scale image classification datasets and network architectures demonstrate the superiority of FedZGE in terms of data heterogeneity, model heterogeneity, communication efficiency, and privacy protection.
\end{abstract}

\begin{links}
	\link{Code}{https://github.com/maxinge8698/FedZGE}
%    \link{Extended version}{https://arxiv.org/abs/XXX}
\end{links}

\section{Introduction}
The remarkable success of deep neural networks is largely attributed to data-driven centralized training schemes, which consolidate data scattered across decentralized clients onto a central server to build high-performing models. However, growing data concerns and regulations impose stringent limitations on data collection and transmission, leading to isolated data silos where clients keep their data locally to ensure privacy and security. To break down this dilemma, federated learning (FL)~\cite{McMahan2017} has emerged as a privacy-preserving decentralized training paradigm that delivers a ``centralized'' model by periodically sharing the parameters of models locally trained by individual clients on their private data with a central server for aggregation. However, the iterative exchange of model parameters between the server and clients exposes FL to several inherent challenges: 1) \emph{Degeneration to data heterogeneity} resulting from the significant divergence between model parameters caused by the non-independent and identically distributed (non-IID) data among clients~\cite{Zhu2021,Ma2022}; 2) \emph{Failure to model heterogeneity} when clients individually design personalized models to meet distinct specifications rather than replicating a homogeneous model architecture for parameter aggregation~\cite{Gao2022,Ye2023}; 3) \emph{Exorbitant communication overhead} that scales up with the number of model parameters hinders its application to larger models or client populations~\cite{Konecny2016,Almanifi2023}; 4) \emph{Vulnerability to privacy attacks} where the shared model parameters can be maliciously exploited to reconstruct clients’ private data in a white-box manner~\cite{Lyu2022,Sharma2024}. 

To tackle these challenges, distillation-based FL~\cite{Li2019,Chang2019,Hu2021a,Itahara2021,Gong2021,Cheng2021,Sattler2022,Gong2022,Ma2023} extends knowledge distillation (KD)~\cite{Hinton2015} to the context of FL by exchanging model outputs associated with a public auxiliary dataset for knowledge transfer, offering a diverse range of benefits such as model-agnostic collaboration, lower communication overhead, and less exposure of sensitive information. However, a desirable auxiliary dataset is not always available since its construction requires careful deliberation and even prior knowledge about clients’ private data to achieve satisfactory performance, which is inconsistent with the privacy-preserving nature of FL.

\begin{table*}[!ht]
	\centering
	\resizebox{\linewidth}{!}{
		\begin{tabular}{ccccccc} 
			\toprule
			\multirow{2.5}{*}{\textbf{Taxonomy}}     & \multirow{2.5}{*}{\textbf{Method }} & \multirow{2.5}{*}{\begin{tabular}[c]{@{}c@{}}\textbf{Auxiliary}\\\textbf{Data }\end{tabular}} & \multirow{2.5}{*}{\begin{tabular}[c]{@{}c@{}}\textbf{Model}\\\textbf{Heterogeneity }\end{tabular}} & \multirow{2.5}{*}{\begin{tabular}[c]{@{}c@{}}\textbf{Black-box}\\\textbf{Access }\end{tabular}} & \multicolumn{2}{c}{\textbf{Exchanged Information }}  \\ 
			\cmidrule{6-7}
			&                                   &                                                                                             &                                                                                                  &                                                                                               & \textbf{Downlink}            & \textbf{Uplink}       \\ 
			\midrule
			parameter-based FL                     & FedAvg                            & -                                                                                           & $\usym{1F5F4}$                                                                                   & $\usym{1F5F4}$                                                                                & model parameter              & model parameter       \\ 
			\midrule
			\multirow{2}{*}{distillation-based FL} & MHAT                              & labeled                                                                                     & $\usym{2714}$                                                                                    & $\usym{2714}$                                                                                 & auxiliary data,~model output & model output          \\
			& DS-FL                             & unlabeled                                                                                   & $\usym{2714}$                                                                                    & $\usym{2714}$                                                                                 & auxiliary data,~model output & model output          \\ 
			\midrule
			\multirow{4}{*}{data-free FL}          & FedGen                            & data-free                                                                                   & $\usym{1F5F4}$                                                                                   & $\usym{1F5F4}$                                                                                & model parameter              & model parameter       \\
			& FedFTG                            & data-free                                                                                   & $\usym{1F5F4}$                                                                                   & $\usym{1F5F4}$                                                                                & model parameter              & model parameter       \\
			& DFRD                              & data-free                                                                                   & $\usym{1F5F4}$                                                                                   & $\usym{1F5F4}$                                                                                & model parameter              & model parameter       \\
			& FedZKT                            & data-free                                                                                   & $\usym{2714}$                                                                                    & $\usym{1F5F4}$                                                                                & model parameter              & model parameter       \\ 
			\midrule
			data-free black-box FL                 & FedZGE                            & data-free                                                                                   & $\usym{2714}$                                                                                    & $\usym{2714}$                                                                                 & synthetic data,~model output & model output          \\
			\bottomrule
		\end{tabular}
	}
	\caption{Comparison of FedZGE with related work.}
	\label{tab:Table_1}
\end{table*}

Instead of assuming a pre-determined auxiliary dataset, data-free FL~\cite{Zhu2021a,Zhang2022,Zhang2022a,Luo2023} eliminates the need for auxiliary data by training a server-side generator to directly synthesize task-specific data samples for knowledge transfer. However, the update rule of the generator depends on the availability of clients’ local models, which involves backpropagating first through local models and then through the generator to yield the gradients. Consequently, it is inevitable for clients to share on-device models with the server for white-box access, which greatly compromises the advantages of distillation-based FL.

To combine the advantages of both distillation-based FL and data-free FL, this motivates us to explore a data-free and black-box FL framework called FedZGE, which leverages Zeroth-order Gradient Estimation~\cite{Nesterov2017} to complete the training of the generator in terms of fidelity, transferability, diversity, and equilibrium by estimating the gradients after flowing through on-device models in a black-box optimization manner, without involving any auxiliary data or sharing any model parameters. In summary, this paper makes the following contributions: 
\begin{itemize}
	\item We propose FedZGE, a data-free black-box FL framework that is granted the ability to launch data-free black-box knowledge transfer across heterogeneous clients via zeroth-order gradient estimation.
	\item Contrary to distillation-based FL approaches that rely on task-relevant auxiliary datasets, FedZGE trains a generator on the server to directly synthesize faithful, transferable, diverse, and balanced data samples for effective knowledge transfer in a data-free manner.
	\item Contrary to data-free FL approaches that require clients to share on-device models, FedZGE estimates the gradients after flowing through these on-device models in a black-box optimization manner to complete the training of the generator, without sharing any model parameters.
	\item Empirical experiments with extensive analysis demonstrate the superiority of FedZGE in terms of data heterogeneity, model heterogeneity, communication efficiency, and privacy protection.
\end{itemize}

\section{Related Work}
\subsection{Parameter-based Federated Learning}
Parameter-based FL represented by FedAvg~\cite{McMahan2017} is a classical FL framework where a central server distributes the global model to individual clients for local training on their respective data, and then the resulting model parameters are sent back for aggregation. Despite its great potential in privacy-sensitive domains~\cite{Liang2021,Liu2022,Li2023}, parameter-based FL still faces challenges such as data heterogeneity, model heterogeneity, communication efficiency, and privacy leakage~\cite{Li2020,Zhang2021,Kairouz2021}. In response to these challenges, numerous variants of FedAvg have emerged~\cite{Tan2022,Alam2022,Bibikar2022,Noble2022}. Unfortunately, they still rely on model parameter exchange and thus lack the ability to address all these issues simultaneously.

\subsection{Distillation-based Federated Learning}
Distillation-based FL~\cite{Mora2022,Wu2023,Qin2024} tackles these challenges by leveraging KD to aggregate local models in a black-box manner. However, it relies heavily on a public auxiliary dataset to synchronize the knowledge produced by these models. Earlier work constructed a task-relevant labeled~\cite{Li2019,Hu2021a,Cheng2021} or unlabeled~\cite{Chang2019,Itahara2021,Sattler2022} dataset with a distribution similar to that of clients' private data as auxiliary data. However, this may explicitly expose clients' private data. Recent efforts~\cite{Gong2021,Gong2022,Ma2023} have relaxed this restriction to a cross-domain unlabeled dataset, whereas finding a desirable auxiliary dataset remains challenging in practice.

\subsection{Data-Free Federated Learning}
Inspired by the recent success of data-free KD~\cite{Lopes2017,Fang2019,Micaelli2019,Fang2022}, data-free FL equips the server with a generator to directly synthesize the data samples necessary for knowledge transfer. Most existing approaches~\cite{Zhu2021a,Zhang2022,Luo2023} are essentially fine-tuning methods that train the generator using the ensemble of local models to further improve the performance of the preliminary global model aggregated from FedAvg. To accommodate heterogeneous clients, recent work~\cite{Zhang2022a} adopts them as data-free methods that transfer knowledge from local models to the global model without parameter aggregation. However, sharing clients' local models with the server is still essential in these frameworks since the training of the generator depends on the availability of local models, which contradicts the benefits of distillation-based FL. A detailed comparison of FedZGE with related work is provided in Table~\ref{tab:Table_1}.

\section{Methodology}
\subsection{Problem Statement}
Consider a FL setting consisting of a central server and $K$ decentralized clients. Each client $k \in \{ 1,2,...,K\} $ locally hosts a private dataset ${{\cal D}_k} = \{ (x_i^{k},y_i^{k})\} _{i = 1}^{{N_k}}$ and an on-device model ${f_k}$ parameterized by ${\theta _k}$, where $x_i^{k} \in {\mathbb{R}^d}$ and $y_i^{k} \in \{ 1,2,...,{\cal C}\} $ are the $i$-th sample and its corresponding label, ${\cal C}$ is the number of classes, and ${N_k}$ is the local sample size. The goal is to learn a global model $f$ parameterized by $\theta$ on the server through multiple rounds of collaborative training across clients while keeping their private data decentralized.

To overcome the limitations imposed by model parameter exchange in FL, distillation-based FL resorts to a publicly available auxiliary dataset ${{\cal D}_{p}}=\{ x_i^{p}\} _{i = 1}^{{N_{p}}}$ for black-box model aggregation via KD. Specifically, in each communication round, the server selects a random set of available clients ${{\cal K}} \subseteq \{ 1,2,...,K\}$ based on a sampling fraction $\varepsilon$ to participate in the FL training, where $|{{\cal K}}| = \left\lceil {\varepsilon  \cdot K} \right\rceil$. Each participating client $k \in {\cal K}$ performs local training to update its local model $f_k$ based on its own private data ${{\cal D}_{k}}$ via the following local optimization objective: 
\begin{equation}\label{eq:Equation_1}
	{{\cal L}_k}({{\cal D}_k};{\theta _k}) = \mathop \mathbb{E}\limits_{(x,y)\sim{{\cal D}_k}} \left[ {{\rm{CE}}({f_k}(x;{\theta _k}),y)} \right],
\end{equation}
where ${\rm{CE}}( \cdot )$ is the cross-entropy loss used to measure training error for a classification task. Subsequently, it downloads the auxiliary data $x_{p}\sim{{\cal D}_{p}}$ from the server and sends back the output ${f_k}({x_{p}};{\theta _k})$ of its local model $f_k$ on $x_p$. The server then aggregates these local outputs into the ensemble output $\sum\nolimits_{k \in {\cal K}} {\frac{{{N_k}}}{N}{f_k}({x_{p}};{\theta _k})}$ through weighted averaging, which assembles knowledge from other peer clients and thus can be used to distill the global model to preserve the knowledge of local models via the following global distillation objective:
\begin{equation}\label{eq:Equation_2}
	\resizebox{0.9\linewidth}{!}{
		${{\cal L}_f}({{\cal D}_{p}};\theta ) = \mathop {\mathbb{E}}\limits_{x_p\sim{{\cal D}_{p}}} \left[ {{\rm{KL(}}\sigma (\sum\limits_{k \in {\cal K}} {\frac{{{N_k}}}{N}{f_k}(x_p;{\theta _k})} ;\tau ){\rm{||}}\sigma (f(x_p;\theta );\tau ){\rm{)}}} \right],$
	}
\end{equation}
where $N = \sum\nolimits_{k \in {\cal K}} {{N_k}}$ is the cumulative sample size, $\sigma ( \cdot ;\tau {\rm{)}}$ is a softmax function with temperature $\tau $ to control the distributional smoothness, and ${\rm{KL}}( \cdot || \cdot )$ is the Kullback-Leibler divergence~\cite{Kullback1951} used to measure the discrepancy between the global model and the ensemble of local models. However, a task-relevant auxiliary dataset may not always be available in practice.

Instead, data-free FL introduces a conditional generator $G$ parameterized by ${\theta _G}$ on the server to capture the training space of clients’ local models in a data-free manner. Specifically, taking a $d_z$-dimensional random noise $z$ sampled from a standard Gaussian distribution ${\cal N}(0,{\bf{1}})$ as the input and a random label $\hat y$ sampled from a uniform distribution ${\cal U}(1,{\cal C})$ as the conditional context, the conditional generator $G$ is trained to synthesize a categorical data $\hat x = G(z,\hat y;{\theta _G})$ by minimizing the cross-entropy loss between the ensemble output and the random label, as follows: 
\begin{equation}\label{eq:Equation_3}
	\resizebox{0.9\linewidth}{!}{
		${\cal L}_G^{fid}(\hat x;{\theta _G}) = \mathop \mathbb{E}\limits_{\scriptstyle\hat x\sim G(z,\hat y;{\theta _G})\hfill\atop
			\scriptstyle\hat y\sim{\cal U}(1,{\cal C})\hfill} \left[ {{\rm{CE(}}\sum\limits_{k \in {\cal K}} {\frac{{{N_k}}}{N}{f_k}(\hat x;{\theta _k})} ,\hat y{\rm{)}}} \right],$
	}
\end{equation}
which allows $\hat x$ to be categorized to one specific class $\hat y$ with a high probability, thereby approximating clients’ private data under label $\hat y$ from a global perspective. Meanwhile, the generated synthetic data sample $\hat x$ is used to transfer knowledge from local models to the global model, as follows:
\begin{equation}\label{eq:Equation_4}
	\resizebox{0.9\linewidth}{!}{
		${{\cal L}_f}(\hat x;\theta ) = \mathop\mathbb{E}\limits_{\hat x\sim G(z,\hat y;{\theta _G})} \left[ {{\rm{KL(}}\sigma (\sum\limits_{k \in {\cal K}} {\frac{{{N_k}}}{N}{f_k}(\hat x;{\theta _k})} ;\tau ){\rm{||}}\sigma (f(\hat x;\theta );\tau ){\rm{)}}} \right].$
	}
\end{equation}
The generator and the global model are alternately trained on the server. However, the update rule of the generator depends on the availability of local models and requires clients to upload their local models to the server for white-box access, which greatly compromises the advantages of distillation-based FL. Our proposed FedZGE aims to address this issue. 

\subsection{Reconstruction of Generator Training}
Although the training objective in Eq.~\eqref{eq:Equation_3} ensures the fidelity of the generated synthetic data samples, they may not be sufficiently transferable, diverse, and balanced due to the lack of explicit intervention in the training process of the generator. Considering that less useful information is transferred from model outputs than from model parameters, it is imperative to improve the quality of generated synthetic data samples so that the global model is able to capture more useful information from local models to better exploit the unique advantages of KD. Therefore, we systematically reconstruct the training of the generator in terms of fidelity, transferability, diversity, and equilibrium.
\paragraph{Improvement of Transferability.} To effectively transfer knowledge from local models to the global model, one solution is to improve the transferability of synthetic data samples. Enlightened by the success of generative adversarial networks (GANs)~\cite{Goodfellow2014,Mirza2014,Radford2015}, we introduce an adversarial loss to encourage the generator $G$ to generate difficult data samples for the training of the global model by maximizing the disagreement between the global model $f$ and the ensemble of local models ${\{ {f_k}\} _{k \in {\cal K}}}$, as follows: 
\begin{equation}\label{eq:Equation_5}
	\resizebox{0.9\linewidth}{!}{
		${\cal L}_G^{adv}(\hat x;{\theta _G}) = \mathop\mathbb{E}\limits_{\hat x\sim G(z,\hat y;{\theta _G})} \left[ { - {\rm{KL}}(\sigma (\sum\limits_{k \in {\cal K}} {\frac{{{N_k}}}{N}{f_k}(\hat x;{\theta _k});\tau } )||\sigma (f(\hat x;\theta );\tau ))} \right].$
	}
\end{equation}
Meanwhile, the generated synthetic data samples need to ensure knowledge consistency between the global model and local models by minimizing the disagreement, as shown in Eq.~\eqref{eq:Equation_4}. As a result, the goals of $G$ and $f$ are to maximize and minimize the disagreement between $f$ and ${\{ {f_k}\} _{k \in {\cal K}}}$, respectively, with the following adversarial game: 
\begin{equation}\label{eq:Equation_6}
	\resizebox{0.9\linewidth}{!}{
		$\mathop {\min }\limits_\theta  \mathop {\max }\limits_{{\theta _G}} \mathop\mathbb{E}\limits_{\hat x\sim G(z,\hat y;{\theta _G})} \left[ {{\rm{KL(}}\sigma {\rm{(}}\sum\limits_{k \in {\cal K}} {\frac{{{N_k}}}{N}{f_k}(\hat x;{\theta _k});\tau )} ||\sigma (f(\hat x;\theta );\tau ){\rm{)}}} \right].$
	}
\end{equation}

\begin{figure*}[!t]
	\centering
	\includegraphics[scale=0.536]{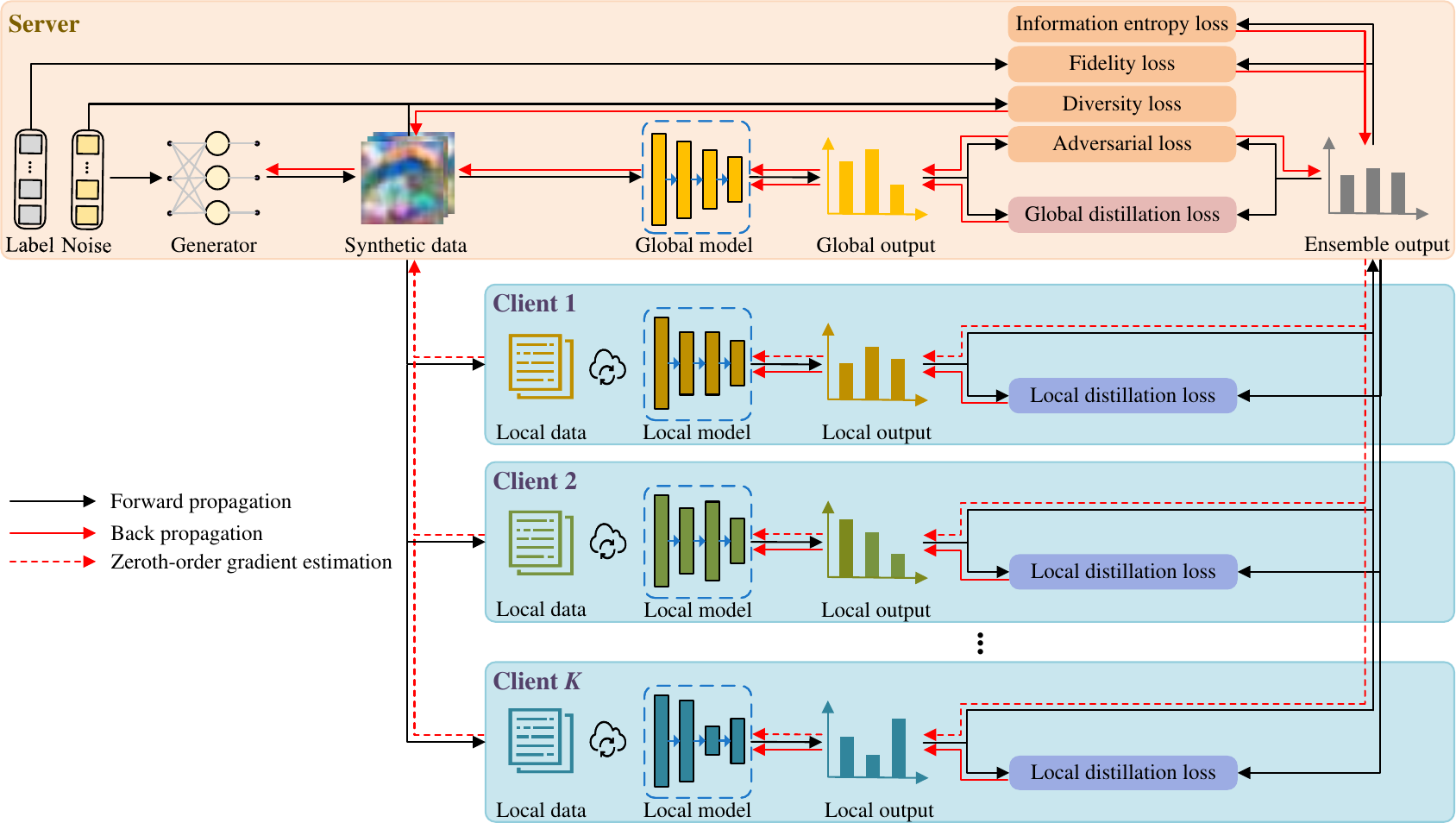}
	\caption{Overview of the proposed FedZGE framework.} 
	\label{fig:Figure_1} 
\end{figure*}

\paragraph{Promotion of Diversity.} The conditional generator $G$ may experience the mode collapse problem~\cite{Mao2019} during training since it tends to learn from the conditional context that determines the main content while ignoring the latent noise that induces diverse outputs, resulting in data samples that are often generated from only a single or a few modes of the distribution. To mitigate this problem, we add a diversity loss to the generator training to promote the diversity and dispersion of the synthetic data samples $\{ {\hat x_i}\} _{i = 1}^B$ generated in each batch over the data space, as follows:
\begin{equation}\label{eq:Equation_7}
	\resizebox{0.9\linewidth}{!}{
		${\cal L}_G^{div}(\hat x;{\theta _G}) = \mathop\mathbb{E}\limits_{\hat x\sim G(z,\hat y;{\theta _G})} \left[ {{e^{\frac{1}{{B*B}}\sum\limits_{i,j \in \{ 1,2,...,B\} } { - ||{{\hat x}_i} - {{\hat x}_j}|{|_2}*||{z_i} - {z_j}|{|_2}} }}} \right],$
	}
\end{equation}
where $B$ is the batch size. This explicitly maximizes the distance between the generated data and the corresponding latent noises to increase the chances of generating data from different modes and thus synthesizing diverse data samples. 

\paragraph{Equilibrium of Distribution.} Uniformly sampling of random labels alone does not guarantee that the generated synthetic data samples are balanced since the overall predicted probability of ensemble outputs is higher on some classes and lower on the others~\cite{Chen2019,Zhang2023}. To this end, we train the generator to maximize the information entropy loss between the synthetic data samples generated in each batch, inducing them to be uniformly distributed across classes. Specifically, we calculate the frequency distribution of these synthetic data samples $\{ {\hat x_i}\} _{i = 1}^B$ belonging to each class $c \in \{ 1,2,...,{\cal C}\}$ from their corresponding ensemble outputs $\{ \sum\nolimits_{k \in {\cal K}} {\frac{{{N_k}}}{N}{f_k}({{\hat x}_i};{\theta _k})} \} _{i = 1}^B$, as follows:
\begin{equation}\label{eq:Equation_8}
	{\bf{p}} = \frac{1}{B}\sum\limits_{i = 1}^B {\sigma (\sum\limits_{k \in {\cal K}} {\frac{{{N_k}}}{N}{f_k}({{\hat x}_i};{\theta _k})} )} ,
\end{equation}
where $\sigma ( \cdot )$ is a standard softmax function. The resulting probability distribution ${\bf{p}} = [{p_1},{p_2},...,{p_{\cal C}}] \in {\mathbb{R}^{\cal C}}$ symbolizes the probability that each batch of synthetic data samples will be generated on each class. By maximizing the information entropy loss, as follows:
\begin{equation}\label{eq:Equation_9}
	{\cal L}_G^{info}(\hat x;{\theta _G}) =  - H({\bf{p}}) = \sum\limits_{c = 1}^{\cal C} {{p_c}} \log ({p_c}),
\end{equation}
each element ${p_c}$ in ${\bf{p}}$ will be equal to $\frac{1}{{\cal C}}$ when the loss reaches the maximum, which implies that the generator can generate data from each class with approximately the same probability, resulting in a balanced batch of synthetic data samples. 

\paragraph{Overall Training Objective.} By combining the above three losses with the original training loss (\emph{i.e.}, fidelity loss ${\cal L}_G^{fid}(\hat x;{\theta _G})$  in Eq.~\eqref{eq:Equation_3}), the generator can be encouraged to synthesize faithful, transferable, diverse, and balanced data samples for effective knowledge transfer, as follows:
\begin{equation}\label{eq:Equation_10}
	{{\cal L}_G}(\hat x;{\theta _G}) = {\cal L}_G^{fid} + {\beta _1}{\cal L}_G^{adv} + {\beta _2}{\cal L}_G^{div} + {\beta _3}{\cal L}_G^{info},
\end{equation}
where ${\beta _1}$, ${\beta _2}$, and ${\beta _3}$ are the tunable scaling factors.

\subsection{Generator Training via Zeroth-Order Gradient Estimation}
As shown in Figure~\ref{fig:Figure_1}, ${{\cal L}_f}(\hat x;\theta )$ in Eq.~\eqref{eq:Equation_4} and ${{\cal L}_G}(\hat x;{\theta _G})$ in Eq.~\eqref{eq:Equation_10} are two loss functions that we wish to minimize over the $d_f$- and $d_G$-dimensional optimization variables $\theta  \in {\mathbb{R}^{{d_f}}}$ and ${\theta _G} \in {\mathbb{R}^{{d_G}}}$, respectively. Accordingly, training the generator $G$ requires using gradient descent to minimize the loss function ${{\cal L}_G}(\hat x;{\theta _G})$ with respect to $\theta_G$, as follows:
\begin{equation}\label{eq:Equation_11}
	{\theta _G} \leftarrow {\theta _G} - {\eta _G}{\nabla _{{\theta _G}}}{{\cal L}_G},
\end{equation}
where ${\eta _G}$ is the learning rate of the generator. Similarly, training the global model $f$ requires minimizing the loss function ${{\cal L}_f}(\hat x;\theta )$ with respect to $\theta$, as follows:
\begin{equation}\label{eq:Equation_12}
	\theta  \leftarrow \theta  - \eta {\nabla _\theta }{{\cal L}_f},
\end{equation}
where $\eta $ is the learning rate of the global model. In the global model training, it is natural to access the parameters of the global model on the server for backpropagation. However, in the generator training, according to the chain rule, it requires backpropagating through local models on individual clients to compute the derivatives of the loss function ${{\cal L}_G}(\hat x;{\theta _G})$ with respect to $\theta_G$, as follows:
\begin{equation}\label{eq:Equation_13}
	\resizebox{0.88\linewidth}{!}{
		${\nabla _{{\theta _G}}}{{\cal L}_G} = \frac{{\partial {{\cal L}_G}}}{{\partial {\theta _G}}} = \frac{{\partial {{\cal L}_G}}}{{\partial (\sum\limits_{k \in {\cal K}} {\frac{{{N_k}}}{N}{f_k}(\hat x;{\theta _k}))} }} \times \frac{{\partial (\sum\limits_{k \in {\cal K}} {\frac{{{N_k}}}{N}{f_k}(\hat x;{\theta _k}))} }}{{\partial {\theta _G}}}.$
	}
\end{equation}
Existing data-free FL approaches assume that clients must share their local models with the server for white-box access to perform the gradient-based first-order optimization for the generator training. However, it is not desirable due to communication burdens and privacy concerns. Instead, treating these local models as on-device black boxes and sharing only their model outputs is consistent with distillation-based FL. Unfortunately, this prevents the computation of first-order gradients for backpropagation. To bridge this gap, we complete the generator training from the perspective of black-box optimization via zeroth-order gradient estimation~\cite{Nesterov2017}.

However, directly estimating the gradients of the loss function ${{\cal L}_G}(\hat x;{\theta _G})$ with respect to the generator parameters ${\theta _G} \in {\mathbb{R}^{{d_G}}}$ is expensive and inefficient considering millions of parameters accompany the generator~\cite{Liu2018}. To overcome the dimensionality curse, we instead choose to estimate the gradients of ${{\cal L}_G}(\hat x;{\theta _G})$ with respect to the synthetic data $\hat x \in {\mathbb{R}^d}$ produced by the generator $G$ since it has a significantly lower dimensionality and is independent of the scaling of the generator, \emph{i.e.}, $d \ll {d_G}$, as follows:
\begin{equation}\label{eq:Equation_14}
	{\nabla _{{\theta _G}}}{{\cal L}_G} = \frac{{\partial {{\cal L}_G}}}{{\partial {\theta _G}}} = \frac{{\partial {{\cal L}_G}}}{{\partial \hat x}} \times \frac{{\partial \hat x}}{{\partial {\theta _G}}},
\end{equation}
where the second term can be directly computed by performing backpropagation through $G$, as follows:
\begin{equation}\label{eq:Equation_15}
	\frac{{\partial \hat x}}{{\partial {\theta _G}}} = \frac{{\partial G(z,\hat y;{\theta _G})}}{{\partial {\theta _G}}}.
\end{equation}
However, the computation of the first term still requires access to the parameters of local models, as follows:
\begin{equation}\label{eq:Equation_16}
	\resizebox{0.88\linewidth}{!}{
		$\frac{{\partial {{\cal L}_G}}}{{\partial \hat x}} = \frac{{\partial {{\cal L}_G}}}{{\partial (\sum\limits_{k \in {\cal K}} {\frac{{{N_k}}}{N}{f_k}(\hat x;{\theta _k}))} }} \times \frac{{\partial (\sum\limits_{k \in {\cal K}} {\frac{{{N_k}}}{N}{f_k}(\hat x;{\theta _k}))} }}{{\partial \hat x}}.$
	}
\end{equation}
To solve Eq.~\eqref{eq:Equation_16}, we leverage zeroth-order gradient estimation that interacts with the loss function ${{\cal L}_G}(\hat x;{\theta _G})$ by simply sending inputs along a set of random direction variables (\emph{i.e.}, $\hat x$ and $\{ \hat x + \epsilon{u_i}\} _{i = 1}^q$) to on-device local models and receiving the corresponding model outputs (\emph{i.e.}, $\sum\nolimits_{k \in {\cal K}} {\frac{{{N_k}}}{N}{f_k}(\hat x;{\theta _k})}$ and $\{ \sum\nolimits_{k \in {\cal K}} {\frac{{{N_k}}}{N}{f_k}(\hat x + \epsilon{u_i};{\theta _k})} \} _{i = 1}^q$) to approximate its first-order gradients with respect to the synthetic data $\hat x$ through finite differences, as follows:
\begin{equation}\label{eq:Equation_17}
	\resizebox{0.88\linewidth} {!}{
		$\frac{{\partial {{\cal L}_G}}}{{\partial \hat x}} \approx {\tilde \nabla _{\hat x}}{{\cal L}_G} = \frac{1}{q}\sum\limits_{i = 1}^q {\frac{{d({{\cal L}_G}(\hat x + \epsilon{u_i};{\theta _G}) - {{\cal L}_G}(\hat x;{\theta _G}))}}{\epsilon}{u_i}},$
	}
\end{equation}
where $d$ is the dimensionality of the synthetic data $\hat x$, $u_i$ is a randomized perturbation variable drawn from a standard Gaussian distribution ${\cal N}(0,{\bf{1}})$, $\epsilon > 0$ is a smoothing parameter that controls the perturbation size, $q$ is the number of random perturbation directions used to acquire finite differences. Consequently, we can compute a biased estimation ${\tilde \nabla _{{\theta _G}}}{{\cal L}_G}$ as an approximation for the true gradients ${\nabla _{{\theta _G}}}{{\cal L}_G}$, as follows:
\begin{equation}\label{eq:Equation_18}
	{\nabla _{{\theta _G}}}{{\cal L}_G} \approx {\tilde \nabla _{{\theta _G}}}{{\cal L}_G} = {\tilde \nabla _{\hat x}}{{\cal L}_G} \times \frac{{\partial G(z,\hat y;{\theta _G})}}{{\partial {\theta _G}}},
\end{equation}
which can then be used to complete the update of the generator through gradient descent, as follows: 
\begin{equation}\label{eq:Equation_19}
	{\theta _G} \leftarrow {\theta _G} - {\eta _G}{\tilde \nabla _{{\theta _G}}}{{\cal L}_G}.
\end{equation}
As a result, the prerequisite of sharing model parameters between the server and clients is eliminated by simply downloading the synthetic data and its perturbations from the server to clients and uploading the corresponding model outputs from clients to the server. It is worth noting that clients only need to perform forward propagation in this process, and thus no excessive computational burden is imposed.

\subsection{Mitigation of Data Heterogeneity}
Only refining the global model may not adequately address the challenge of data heterogeneity since the non-IID data distribution among clients is not mitigated. To this end, we leverage the balanced synthetic data samples and the corresponding ensemble outputs as additional supervision for local distillation, aiming to smooth the non-IID data distributions among local models and narrow their representation gaps towards global generalization. Specifically, after transferring knowledge from local models to the global model on the server, the ensemble output also needs to be transmitted back to clients, with which each client can further regulate its local model via the following local distillation objective:
\begin{equation}\label{eq:Equation_20}
	\resizebox{0.88\linewidth}{!}{
		${{\cal L}'_k}(\hat x;{\theta _k}) = \mathop\mathbb{E}\limits_{\hat x\sim G(z,\hat y;{\theta _G})} \left[ {{\rm{KL(}}\sigma (\sum\limits_{k \in {\cal K}} {\frac{{{N_k}}}{N}{f_k}(\hat x;{\theta _k})} ;\tau ){\rm{||}}\sigma ({f_k}(\hat x;{\theta _k});\tau ){\rm{)}}} \right]{\rm{.}}$
	}
\end{equation}
In this way, the central server can coordinate decentralized clients to collaboratively train a global model by aggregating their distributed knowledge without involving auxiliary data or sharing on-device models. See Appendix A for detailed algorithmic procedures of the proposed FedZGE framework.	

\begin{table*}[!ht]
	\centering
	\resizebox{\linewidth}{!}{
		\begin{tabular}{cccccccccc} 
			\toprule
			\multirow{2.5}{*}{\textbf{Method}} & \multicolumn{3}{c}{\textbf{ResNet-18}}                     & \multicolumn{3}{c}{\textbf{ResNet-34}}                     & \multicolumn{3}{c}{\textbf{ResNet-50}}                      \\ 
			\cmidrule{2-10}
			& $\alpha$=1          & $\alpha$=0.1        & Comm           & $\alpha$=1          & $\alpha$=0.1        & Comm           & $\alpha$=1          & $\alpha$=0.1        & Comm            \\ 
			\midrule
			\multicolumn{1}{l}{FedAvg}       & 80.42$\pm$0.34          & 68.90$\pm$0.51          & 83.31          & 81.36$\pm$0.19          & 69.48$\pm$0.25          & 158.62         & 82.01$\pm$0.16          & 70.12$\pm$0.41          & 175.30          \\ 
			\midrule
			MHAT                             & 77.66$\pm$0.18          & 65.17$\pm$0.36          & 5.76           & 78.29$\pm$0.24          & 66.06$\pm$0.36          & 5.76           & 78.83$\pm$0.09          & 66.53$\pm$0.19          & 5.76            \\
			DS-FL                            & 77.35$\pm$0.23          & 64.75$\pm$0.29          & 5.76           & 78.06$\pm$0.33          & 65.83$\pm$0.60          & 5.76           & 78.57$\pm$0.30          & 66.30$\pm$0.27          & 5.76            \\ 
			\midrule
			FedGen                           & 81.29$\pm$0.31          & 69.97$\pm$0.40          & 90.92          & 82.33$\pm$0.12          & 71.48$\pm$0.43          & 166.23         & 82.96$\pm$0.17          & 72.10$\pm$0.27          & 182.91          \\
			FedFTG                           & 81.44$\pm$0.17          & 70.37$\pm$0.37          & 83.31          & 82.93$\pm$0.30          & 72.07$\pm$0.19          & 158.62         & 83.38$\pm$0.20          & 72.50$\pm$0.54          & 175.30          \\
			DFRD                             & 81.78$\pm$0.11          & 70.86$\pm$0.24          & 83.31          & 83.10$\pm$0.22          & 72.31$\pm$0.25          & 158.62         & 83.59$\pm$0.12          & 72.72$\pm$0.36          & 175.30          \\
			FedZKT                           & 81.20$\pm$0.16          & 70.06$\pm$0.32          & 83.31          & 82.64$\pm$0.17          & 71.78$\pm$0.24          & 158.62         & 83.21$\pm$0.31          & 72.28$\pm$0.51          & 175.30          \\ 
			\midrule
			FedZGE                           & \textbf{82.19$\pm$0.20} & \textbf{71.35$\pm$0.28} & \textbf{63.17} & \textbf{83.40$\pm$0.12} & \textbf{72.65$\pm$0.18} & \textbf{63.17} & \textbf{83.78$\pm$0.21} & \textbf{72.94$\pm$0.25} & \textbf{63.17}  \\
			\bottomrule
		\end{tabular}
	}
	\caption{Test accuracy (\%) and communication overhead (GB) on the CIFAR-10 dataset under the homogeneous FL setting.}
	\label{tab:Table_2}
\end{table*}

\begin{table*}[!ht]
	\centering
	\resizebox{\linewidth}{!}{
		\begin{tabular}{cccccccccc} 
			\toprule
			\multirow{2.5}{*}{\textbf{Method}} & \multicolumn{3}{c}{\textbf{ResNet-18}}                     & \multicolumn{3}{c}{\textbf{ResNet-34}}                     & \multicolumn{3}{c}{\textbf{ResNet-50}}                      \\ 
			\cmidrule{2-10}
			& $\alpha$=1          & $\alpha$=0.1        & Comm           & $\alpha$=1          & $\alpha$=0.1        & Comm           & $\alpha$=1          & $\alpha$=0.1        & Comm            \\ 
			\midrule
			\multicolumn{1}{l}{FedAvg}       & 56.87$\pm$0.40          & 41.63$\pm$0.55          & 83.65          & 58.24$\pm$0.35          & 42.11$\pm$0.43          & 158.97         & 58.89$\pm$0.16           & 43.04$\pm$0.50          & 176.68          \\ 
			\midrule
			MHAT                             & 53.12$\pm$0.31          & 38.71$\pm$0.36          & 6.10           & 54.00$\pm$0.09          & 39.34$\pm$0.22          & 6.10           & 54.76$\pm$0.28          & 40.11$\pm$0.44          & 6.10            \\
			DS-FL                            & 52.71$\pm$0.28          & 38.25$\pm$0.19          & 6.09           & 53.65$\pm$0.13          & 39.08$\pm$0.24          & 6.09           & 54.23$\pm$0.36          & 39.79$\pm$0.37          & 6.09            \\ 
			\midrule
			FedGen                           & 57.42$\pm$0.15          & 42.84$\pm$0.35          & 91.30          & 58.67$\pm$0.27          & 43.45$\pm$0.35          & 166.61         & 59.06$\pm$0.13          & 44.09$\pm$0.29          & 184.32          \\
			FedFTG                           & 57.76$\pm$0.16          & 43.03$\pm$0.41          & 83.65          & 58.86$\pm$0.16          & 43.74$\pm$0.19          & 158.97         & 59.54$\pm$0.17          & 44.47$\pm$0.18          & 176.68          \\
			DFRD                             & 58.00$\pm$0.25          & 43.20$\pm$0.21          & 83.65          & 59.39$\pm$0.23          & 43.96$\pm$0.26          & 158.97         & 59.93$\pm$0.26          & 44.51$\pm$0.33          & 176.68          \\
			FedZKT                           & 57.58$\pm$0.32          & 42.93$\pm$0.14          & 83.65          & 58.50$\pm$0.14          & 43.59$\pm$0.31          & 158.97         & 58.74$\pm$0.19          & 44.16$\pm$0.51          & 176.68          \\ 
			\midrule
			FedZGE                           & \textbf{58.41$\pm$0.11} & \textbf{43.66$\pm$0.23} & \textbf{65.18} & \textbf{59.67$\pm$0.25} & \textbf{44.24$\pm$0.32} & \textbf{65.18} & \textbf{60.02$\pm$0.17} & \textbf{44.67$\pm$0.36} & \textbf{65.18}  \\
			\bottomrule
		\end{tabular}
	}
	\caption{Test accuracy (\%) and communication overhead (GB) on the CIFAR-100 dataset under the homogeneous FL setting.}
	\label{tab:Table_3}
\end{table*}

\section{Experiments}
\subsection{Experimental Setup}
\paragraph{Datasets.} The evaluation is conducted on two image classification datasets commonly used in FL research: 1) \textbf{CIFAR-10}~\cite{Krizhevsky2009}; 2) \textbf{CIFAR-100}~\cite{Krizhevsky2009}. To simulate data heterogeneity among clients, we follow prior work~\cite{Hsu2019} to heterogeneously partition the training set of each dataset among clients using a Dirichlet distribution $Dir(\alpha)$, where $\alpha$ is a concentration parameter that controls the degree of non-IID, with smaller values indicating more heterogeneous data distribution. To ensure reliable performance evaluation, we run the experiments three times with different random seeds and report the average accuracy with standard deviation of the global model on the original test set. See Appendix B for more dataset statistics and partitioning details.

\paragraph{Baselines.} We comprehensively compare FedZGE with three types of FL frameworks: 1) \textbf{parameter-based FL approaches}, including FedAvg~\cite{McMahan2017}, which aggregates knowledge through iterative model parameter exchange; 2) \textbf{distillation-based FL approaches}, including MHAT~\cite{Hu2021a} and DS-FL~\cite{Itahara2021}, which rely on a labeled and an unlabeled auxiliary dataset, respectively, to iteratively exchange model outputs for knowledge transfer; 3) \textbf{data-free FL approaches}, including FedGen~\cite{Zhu2021a}, FedFTG~\cite{Zhang2022}, DFRD~\cite{Luo2023}, and FedZKT~\cite{Zhang2022a}, which still require clients to share their local models for data-free knowledge transfer.

\begin{table}[!bt]
	\centering
	\scriptsize
	\setlength{\tabcolsep}{1.0mm} % 列间距
	\renewcommand{\arraystretch}{0.7} %行间距
%	\resizebox{\linewidth}{!}{
		\begin{tabular}{ccccccc} 
			\toprule
			\multirow{2.5}{*}{\textbf{Method}} & \multicolumn{3}{c}{\textbf{CIFAR-10}}                      & \multicolumn{3}{c}{\textbf{CIFAR-100}}                      \\ 
			\cmidrule{2-7}
			& $\alpha$=1          & $\alpha$=0.1        & Comm           & $\alpha$=1          & $\alpha$=0.1        & Comm            \\ 
			\midrule
			FedAvg                           & -                   & -                   & -              & -                   & -                   & -               \\ 
			\midrule
			MHAT                             & 78.50$\pm$0.17          & 66.30$\pm$0.28          & 5.76           & 54.33$\pm$0.17          & 39.77$\pm$0.21          & 6.10            \\
			DS-FL                            & 78.23$\pm$0.09          & 65.95$\pm$0.35          & 5.76           & 54.81$\pm$0.31          & 39.63$\pm$0.45          & 6.10            \\ 
			\midrule
			FedGen                           & -                   & -                   & -              & -                   & -                   & -               \\
			FedFTG                           & 83.09$\pm$0.13          & 72.24$\pm$0.24          & 159.00         & 59.31$\pm$0.24          & 44.09$\pm$0.10          & 160.07          \\
			DFRD                             & 83.21$\pm$0.22          & 72.52$\pm$0.11          & 159.00         & 59.52$\pm$0.19          & 44.26$\pm$0.19          & 160.07          \\
			FedZKT                           & 83.00$\pm$0.26          & 71.94$\pm$0.45          & 142.70         & 58.59$\pm$0.11          & 43.90$\pm$0.27          & 143.46          \\ 
			\midrule
			FedZGE                           & \textbf{83.60$\pm$0.14} & \textbf{72.81$\pm$0.33} & \textbf{63.17} & \textbf{59.73$\pm$0.21} & \textbf{44.44$\pm$0.30} & \textbf{65.18}  \\
			\bottomrule
		\end{tabular}
%	}
	\caption{Test accuracy (\%) and communication overhead (GB) on the CIFAR-10 and CIFAR-100 datasets under the heterogeneous FL setting.}
	\label{tab:Table_4}
\end{table}

\paragraph{Settings.} The experiments are performed under two distinct FL settings with the key hyperparameters $\alpha$=\{1, 0.1\}, $K$=10, $\varepsilon$=1, $T$=100, $B$=500, and $q$=10: 1) \textbf{homogeneous FL setting}, where clients are forced to replicate homogeneous local models with the same architecture as the global model. We employ three types of network architectures to explore the effects of model scaling, including ResNet-18, ResNet-34, and ResNet-50~\cite{He2016}; 2) \textbf{heterogeneous FL setting}, where clients are allowed to independently design heterogeneous local models. We employ ResNet-50 as the global model and allocate ResNet-18, ResNet-34, and ResNet-50 as the local models to clients in a ratio of 3:3:4. See Appendix C for full implementation details.

\subsection{Experimental Results}
\paragraph{Robustness to Data Heterogeneity.} Tables~\ref{tab:Table_2} and \ref{tab:Table_3} report the test accuracy and communication overhead of different FL approaches on the CIFAR-10 and CIFAR-100 datasets under the homogeneous FL setting, respectively. A notable observation is that the performance of all FL approaches deteriorates dramatically as the degree of data heterogeneity among clients increases. However, FedZGE significantly outperforms distillation-based FL approaches and achieves comparable performance to data-free FL approaches at different levels of data heterogeneity as well as different scales of network architectures, demonstrating its strong robustness in data heterogeneous FL scenarios.

\paragraph{Compatibility to Model Heterogeneity.} Table~\ref{tab:Table_4} reports the test accuracy and communication overhead of different FL approaches on the CIFAR-10 and CIFAR-100 datasets under the heterogeneous FL setting. We can observe that FedZGE consistently outperforms distillation-based FL approaches with superior performance and overwhelms data-free FL approaches with less communication overhead, without sharing any model parameters, suggesting that FedZGE is more advantageous in model heterogeneous FL scenarios. 

\begin{table}[!ht]
	\centering
	\scriptsize
	\setlength{\tabcolsep}{0.5mm} % 列间距
	\renewcommand{\arraystretch}{0.7} %行间距
%	\resizebox{\linewidth}{!}{
		\begin{tabular}{ccccccc} 
			\toprule
			\multirow{2.5}{*}{\textbf{Method}} & \multicolumn{3}{c}{\textbf{CIFAR-10}}                      & \multicolumn{3}{c}{\textbf{CIFAR-100}}                      \\ 
			\cmidrule{2-7}
			& $\alpha$=1          & $\alpha$=0.1        & Comm           & $\alpha$=1          & $\alpha$=0.1        & Comm            \\ 
			\midrule
			FedZGE                           & \textbf{83.78$\pm$0.21} & \textbf{72.94$\pm$0.25} & \textbf{63.17} & \textbf{60.02$\pm$0.17} & \textbf{44.67$\pm$0.36} & \textbf{65.18}  \\ 
			\midrule
			FedZGE ($q$=1)                   & 83.10$\pm$0.30          & 72.08$\pm$0.22          & 11.50          & 58.81$\pm$0.14          & 43.85$\pm$0.50          & 12.00           \\
			FedZGE ($q$=5)                   & 83.44$\pm$0.26          & 72.33$\pm$0.33          & 34.46          & 59.37$\pm$0.20          & 44.40$\pm$0.41          & 35.64           \\
			FedZGE ($q$=20)                  & 83.88$\pm$0.17          & 73.00$\pm$0.40          & 120.57         & 60.08$\pm$0.07          & 44.76$\pm$0.32          & 124.26          \\ 
			\midrule
			w/~${\cal L}_G^{fid}$                & 82.90$\pm$0.08          & 72.01$\pm$0.13          & 63.17          & 59.04$\pm$0.11          & 43.97$\pm$0.27          & 65.18           \\
			w/o ${\cal L}_G^{adv}$               & 83.54$\pm$0.18          & 72.70$\pm$0.37          & 63.17          & 59.58$\pm$0.14          & 44.39$\pm$0.48          & 65.18           \\
			w/o~${\cal L}_G^{div}$               & 83.70$\pm$0.26          & 72.82$\pm$0.21          & 63.17          & 59.74$\pm$0.25          & 44.58$\pm$0.24          & 65.18           \\
			w/o~${\cal L}_G^{info}$              & 83.58$\pm$0.10          & 72.76$\pm$0.46          & 63.17          & 59.70$\pm$0.18          & 44.44$\pm$0.25          & 65.18           \\ 
			\midrule
			w/o~${\cal L}'_k$                  & 83.47$\pm$0.24          & 72.68$\pm$0.36          & 63.17          & 59.46$\pm$0.28          & 44.30$\pm$0.33          & 65.18           \\
			\bottomrule
		\end{tabular}
%	}
	\caption{Ablation study of different FedZGE variants on the CIFAR-10 and CIFAR-100 datasets using ResNet-50.}
	\label{tab:Table_5}
\end{table}

\paragraph{Alleviation to Communication Overhead.} The formulations to calculate communication overhead for different FL methods are listed in Appendix D. As shown in Tables~\ref{tab:Table_2}, \ref{tab:Table_3} and \ref{tab:Table_4}, the communication overhead of FedZGE is independent of model size, and consistently lies between distillation-based FL approaches and data-free FL approaches. In contrast to data-free FL that sacrifices the benefits of black-box access and communication-efficient collaboration offered by distillation-based FL in exchange for improved performance, FedZGE retains the low communication overhead of distillation-based FL and the high performance of data-free FL simultaneously, serving as a compromise between them to provide a more effective and efficient FL solution.

\paragraph{Resistance to Privacy Leakage.} Distillation-based FL and data-free FL require transmitting real data samples and model parameters between the server and clients, respectively, which may raise privacy-related concerns since a malicious attacker could exploit this information to reconstruct clients' private data. Instead, FedZGE transmits synthetic data samples that capture only high-level feature patterns, which are incomprehensible to humans and express the global characteristics of private data, thereby providing stronger privacy guarantees than previous FL approaches. To illustrate this, we provide a visual comparison between the synthetic data and the private data on the CIFAR-10 and CIFAR-100 datasets in Appendix E. 

\subsection{Ablation Study}
\paragraph{Effects of Loss Function Design.} To explore the contribution of different auxiliary losses to the generator training, we conduct a leave-one-out test to show the results after removing individual losses (denoted as w/o ${\cal L}_G^{adv}$, w/o ${\cal L}_G^{div}$, and w/o ${\cal L}_G^{info}$, respectively) and all losses (denoted as w/ ${\cal L}_G^{fid}$). Moreover, we remove local distillation losses from clients (denoted as w/o ${{\cal L}'_k}$) to explore the role they play. As shown in Table~\ref{tab:Table_5}, it can observed that using only ${\cal L}_G^{fid}$ to train the generator leads to poor performance, and the absence of a single loss leads to significant performance degradation, suggesting that each component contributes to the improvement of the generator. Furthermore, removing local distillation losses leads to more pronounced performance degradation in scenarios with severe data heterogeneity, highlighting its effectiveness in mitigating data heterogeneity. 

\begin{figure}[!t]
	\centering
	\subfigure[CIFAR-10, $\alpha$=1]{
		\label{fig:Figure_2(a)}
		\includegraphics[scale=0.25]{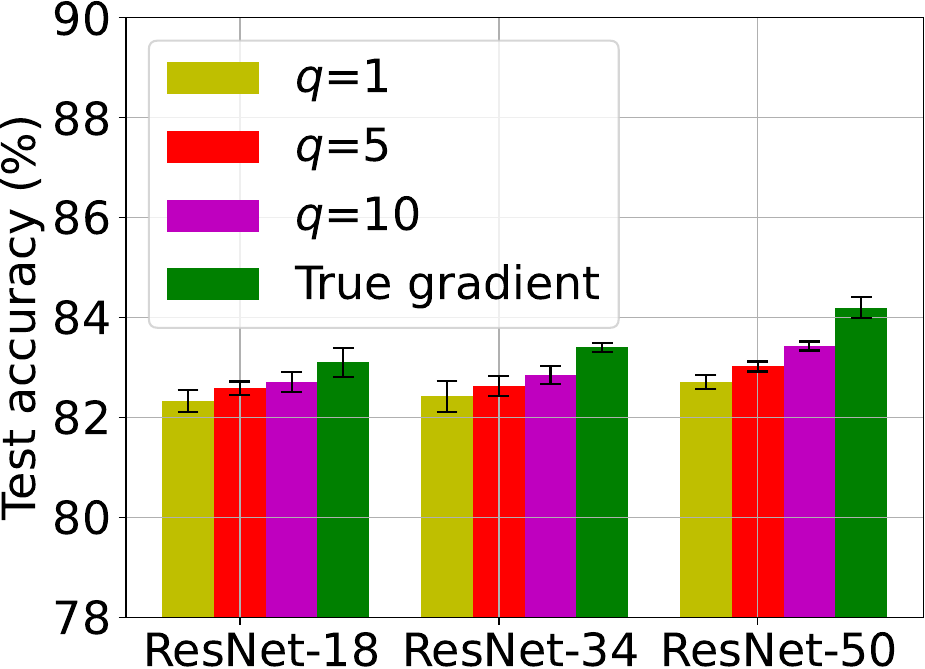}
	}
	\subfigure[CIFAR-10, $\alpha$=0.1]{
		\label{fig:Figure_2(b)}
		\includegraphics[scale=0.25]{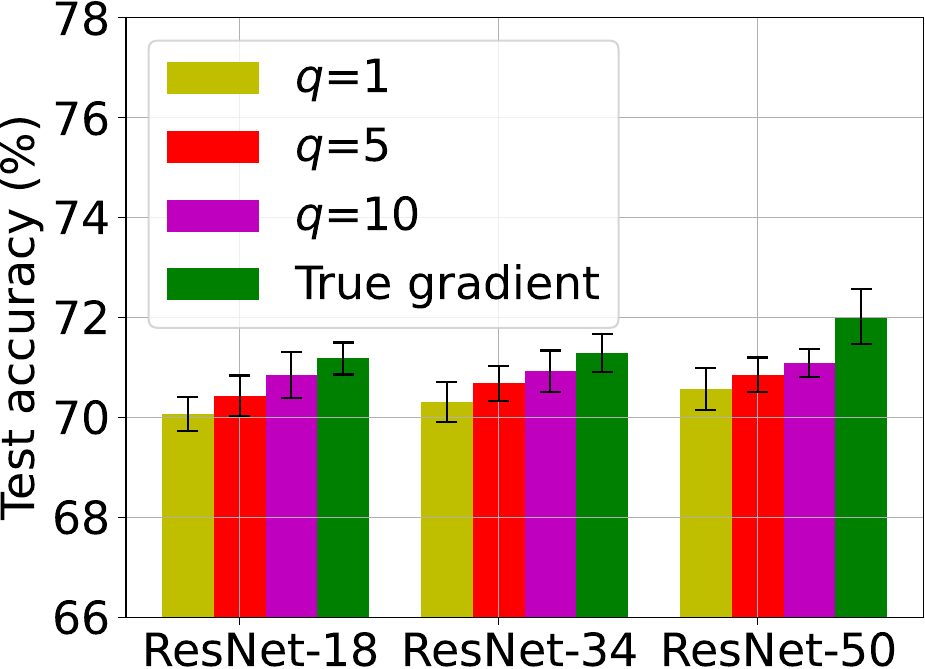}
	}
	\\
	\subfigure[CIFAR-100, $\alpha$=1]{
		\label{fig:Figure_2(c)}
		\includegraphics[scale=0.25]{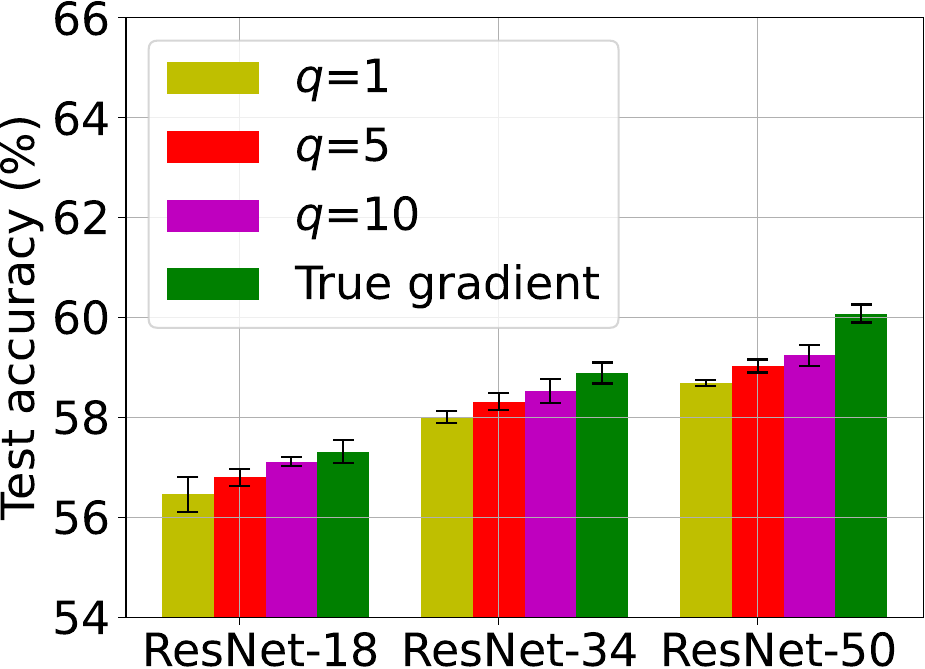}
	}
	\subfigure[CIFAR-100, $\alpha$=0.1]{
		\label{fig:Figure_2(d)}
		\includegraphics[scale=0.25]{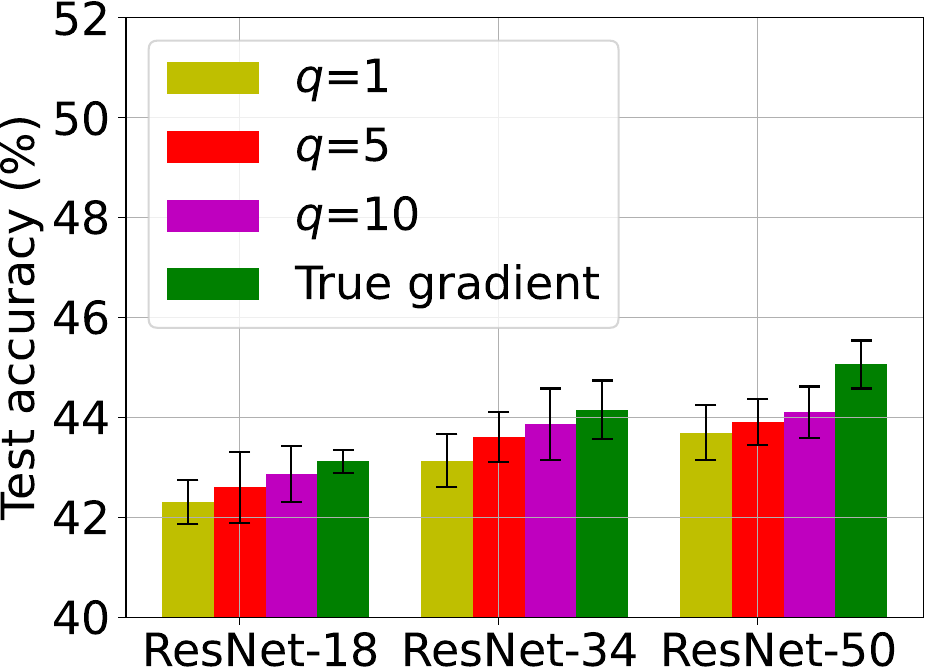}
	}
	\caption{Ablation study of FedZGE and FedZGE with true gradients on the CIFAR-10 and CIFAR-100 datasets.}
	\label{fig:Figure_2}
\end{figure}

\paragraph{Effects of Perturbation Directions.} To examine the effects of the number of random perturbation directions $q$ on FedZGE, we sample different $q$=\{1, 5, 20\} for zeroth-order gradient estimation. As shown in Table~\ref{tab:Table_5}, it can be observed that a larger value of $q$ leads to a more accurate gradient estimation but increases the communication overhead. Therefore, there is a trade-off between test accuracy and communication overhead for FedZGE. Empirically, adopting $q$=10 is the best practice since it performs comparably or even better than other baselines with less communication overhead.

\paragraph{Effects of Gradient Estimation.} To understand how the gradient estimation affects the performance of FedZGE, we modify FedZGE for analytical purposes by allowing clients to upload their local models to the server for white-box access to obtain the true gradient. As shown in Figure~\ref{fig:Figure_2}, It can be observed that the performance of FedZGE is close to that when the true gradient of the generator is available. However, it is closely related to the size of the local models to be estimated due to the higher the complexity of the black-box network architecture, the larger the difference between the estimated gradient and the true gradient. More experimental results and analysis can be found in Appendix F.

\section{Conclusion}
To combine the advantages of both distillation-based FL and data-free FL, this paper introduces FedZGE, a data-free and black-box FL framework that leverages Zeroth-order Gradient Estimation to complete the training of the generator in terms of fidelity, transferability, diversity, and equilibrium, without involving any auxiliary data or sharing any model parameters. Extensive experiments on large-scale image classification datasets and network architectures demonstrate the superiority of FedZGE in terms of data heterogeneity, model heterogeneity, communication efficiency, and privacy protection. Future work will explore the application of data-free black-box FL to more challenging textual domains with discrete input spaces and large foundation models.

\section{Acknowledgements}
This work was supported by the National Natural Science Foundation of China (NSFC) under Grant Nos.61966038 and 62266051, and the Postgraduate Research and Innovation Foundation of Yunnan University under Grant No.KC-23236531. The authors would like to thank the anonymous reviewers for their constructive comments. 

\bibliography{References}

%%%%%%%%%%%%%%%%%%%%%%%%%%%%%%%%%%%%%%%%%%%%%%%%%%%%%%%%%%%%%
\newpage
\appendix

\section{Appendix}
This appendix contains supplementary material that provides additional details and experimental analysis for the main paper, which is organized as follows:
\begin{itemize}
	\item Section A provides the algorithmic procedures.
	\item Section B provides the dataset statistics and partitioning details.
	\item Section C provides the implementation details.
	\item Section D provides the formulations of communication overhead.
	\item Section E provides the visualization of private data and synthetic data.
	\item Section F provides the extended experimental results.
	%		\item Section G provides the limitations of this paper.
\end{itemize}

\subsection{A. Algorithms}
The detailed training procedures of the proposed FedZGE framework is summarized in Algorithm~\ref{alg:Algorithm_1}.
\begin{algorithm}[!ht]
	\caption{Data-Free Black-Box Federated Learning via Zeroth-order Gradient Estimation (FedZGE)}
	\label{alg:Algorithm_1}
	\textbf{Input}: On-device local datasets $\{ {{\cal D}_k}\} _{k = 1}^K$, on-device local models $\{ {f_k}\} _{k = 1}^K$ with parameters $\{ {\theta _k}\} _{k = 1}^K$, global model $f$ with parameters $\theta$, generator $G$ with parameters $\theta_G$, communication rounds $T$, local training epochs $\{ {E _k}\} _{k = 1}^K$, local distillation epochs $\{ {E _k'}\} _{k = 1}^K$, global distillation epochs $E$, learning rate of local models $\{ {\eta _k}\} _{k = 1}^K$, learning rate of global model $\eta$, learning rate of generator $\eta_G$, random perturbation directions $q$.
	\begin{algorithmic}[1] %[1] enables line numbers
		\STATE Initialize parameters $\theta ^0$, $\{ \theta _k^0\} _{k = 1}^K$, and $\theta _G^0$
		\FOR{each communication round $t \in [1,T]$}
		\STATE ${{\cal K}^t} \leftarrow $ a random set of $\left\lceil {\varepsilon  \cdot K} \right\rceil $ clients
		\STATE // \textit{Local Update} 
		\FOR{each client $k \in {{\cal K}^t}$ \textit{in parallel}}
		\FOR{each local training epoch $e \in [1, E_k]$}
		\STATE $\theta _k^t \leftarrow \theta _k^{t - 1} - {\eta _k}{\nabla _{{\theta _k}}}{{\cal L}_k}({{\cal D}_k};\theta _k^{t - 1})${\hfill $\triangleright$Eq.~\eqref{eq:Equation_1}}
		\ENDFOR
		\ENDFOR
		\STATE Sample a batch of random noises $z\sim {\cal N}(0,{\bf{1}})$
		\STATE Sample a batch of random labels $\hat y\sim {\cal U}(1,{\cal C})$
		\STATE Generate a batch of synthetic data $\hat x \leftarrow G(z,\hat y;\theta _G^{t - 1})$
		\STATE Sample a batch of random perturbations $\{ {u_i}\} _{i = 1}^q$ to construct $\{ \hat x + \epsilon{u_i}\} _{i = 1}^q$
		\STATE Server distributes $\hat x$ and $\{ \hat x + \epsilon{u_i}\} _{i = 1}^q$ to clients ${{\cal K}^t}$
		\STATE // \textit{Local Prediction} 
		\FOR{each client $k \in {{\cal K}^t}$ \textit{in parallel}}
		\STATE Compute and upload local outputs ${f_k}(\hat x;\theta _k^t)$ and $\{ {f_k}(\hat x + \epsilon{u_i};\theta _k^t)\} _{i = 1}^q$ to the server
		\ENDFOR
		\STATE // \textit{Server Aggregation} 
		\STATE Server collects and aggregates local outputs into ensemble outputs  $\sum\nolimits_{k \in {{\cal K}^t}} {\frac{{{N_k}}}{N}{f_k}(\hat x;\theta _k^t)}$ and $\{ \sum\nolimits_{k \in {{\cal K}^t}} {\frac{{{N_k}}}{N}{f_k}(\hat x + \epsilon{u_i};\theta _k^t)} \} _{i = 1}^q$
		\STATE // \textit{Generator Update} 
		\STATE Server estimates ${\tilde \nabla _{{\theta _G}}}{{\cal L}_G}(\hat x;\theta _G^{t - 1})${\hfill $\triangleright$Eq.~\eqref{eq:Equation_18}}
		\STATE $\theta _G^t \leftarrow \theta _G^{t - 1} - {\eta _G}{\tilde \nabla _{{\theta _G}}}{{\cal L}_G}(\hat x;\theta _G^{t - 1})${\hfill $\triangleright$Eq.~\eqref{eq:Equation_19}}
		\STATE // \textit{Global Model Update} 
		\FOR{each global distillation epoch $e \in [1, E]$}
		\STATE ${\theta ^t} \leftarrow {\theta ^{t - 1}} - \eta {\nabla _\theta }{{\cal L}_f}(\hat x;{\theta ^{t - 1}})${\hfill $\triangleright$Eq.~\eqref{eq:Equation_12}}
		\ENDFOR
		\STATE Server distributes $\sum\nolimits_{k \in {{\cal K}^t}} {\frac{{{N_k}}}{N}{f_k}({{\hat x}};\theta _k^t)}$  to clients ${{\cal K}^t}$
		\STATE // \textit{Local Distillation} 
		\FOR{each client $k \in {{\cal K}^t}$ \textit{in parallel}}
		\FOR{each local distillation epoch $e \in [1, E_k']$}
		\STATE $\theta _k^t \leftarrow \theta _k^t - {\eta _k}{\nabla _{{\theta _k}}}{{\cal L}'_k}(\hat x;\theta _k^t)${\hfill $\triangleright$Eq.~\eqref{eq:Equation_20}}
		\ENDFOR
		\ENDFOR
		\ENDFOR
		\RETURN{${\theta^T}$}
	\end{algorithmic}
\end{algorithm}

\subsection{B. Datasets}
\paragraph{Dataset Statistics.} Table~\ref{tab:Table_6} provides information about the image size (\#size), number of classes (\#class), number of training samples (\#train), and number of test samples (\#test) of the datasets used for the experiments.

\begin{table}[!b]
	\centering
	\resizebox{\linewidth}{!}{
		\begin{tabular}{ccccc} 
			\toprule
			\textbf{Dataset} & \textbf{\#size}       & \textbf{\#class} & \textbf{\#train} & \textbf{\#test}  \\ 
			\midrule
			CIFAR-10         & 3$\times$32$\times$32 & 10               & 50,000           & 10,000           \\ 
			\midrule
			CIFAR-100        & 3$\times$32$\times$32 & 100              & 50,000           & 10,000           \\
			\bottomrule
		\end{tabular}
	}
	\caption{Statistics of the datasets.}
	\label{tab:Table_6}
\end{table}

\begin{figure*}[!t]
	\centering
	\subfigure[CIFAR-10, $\alpha$=1, seed=0]{
		\label{fig:Figure_3(a)}
		\includegraphics[scale=0.34]{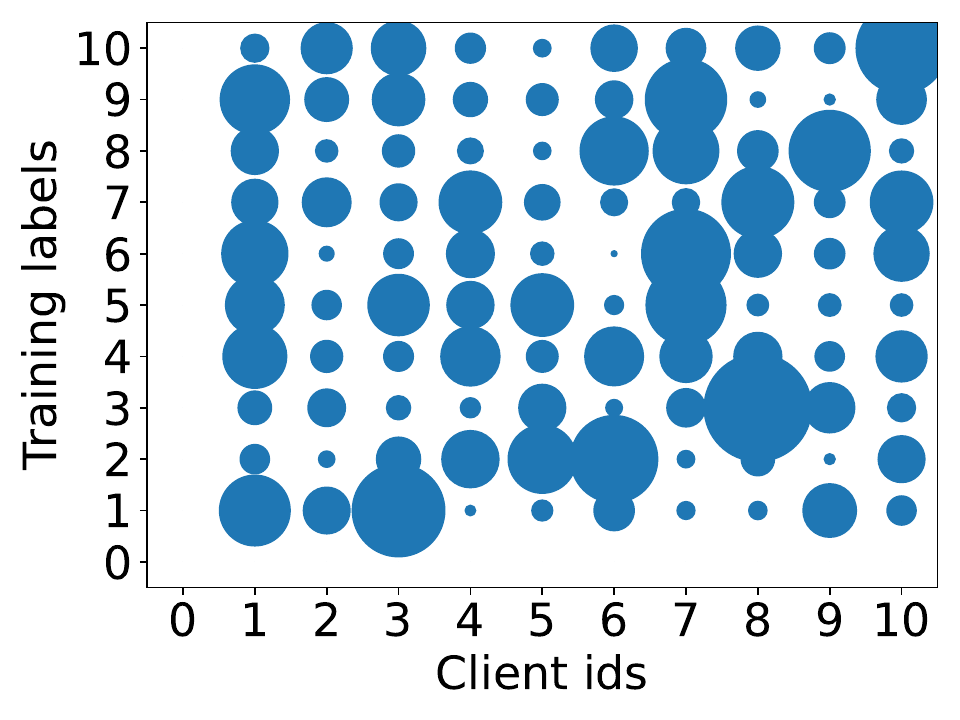}
	}
	\subfigure[CIFAR-10, $\alpha$=1, seed=1]{
		\label{fig:Figure_3(b)}
		\includegraphics[scale=0.34]{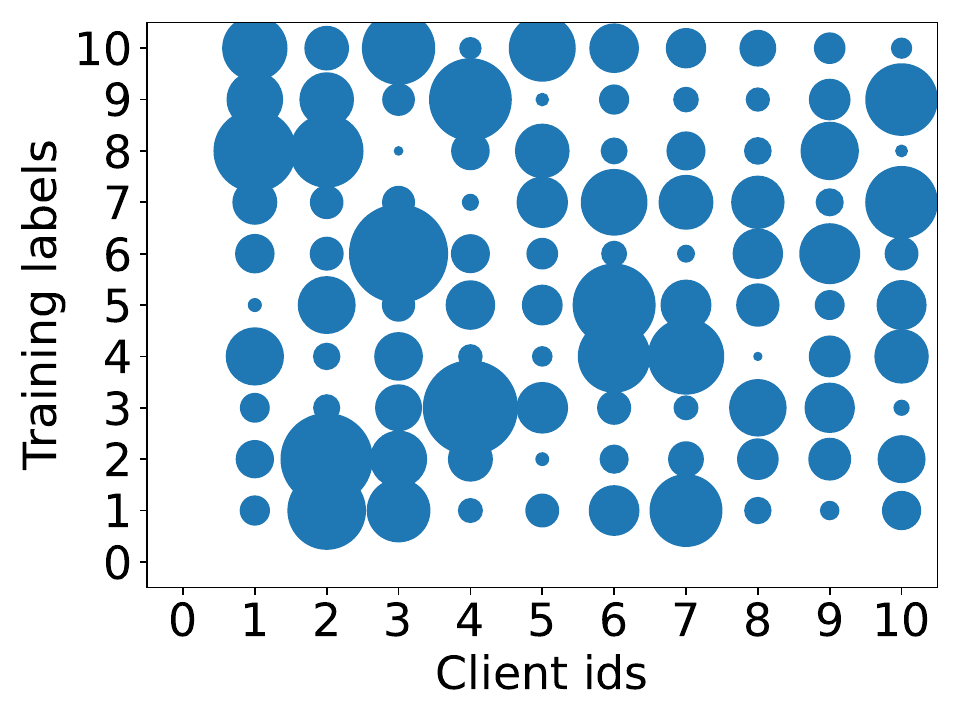}
	}
	\subfigure[CIFAR-10, $\alpha$=1, seed=2]{
		\label{fig:Figure_3(c)}
		\includegraphics[scale=0.34]{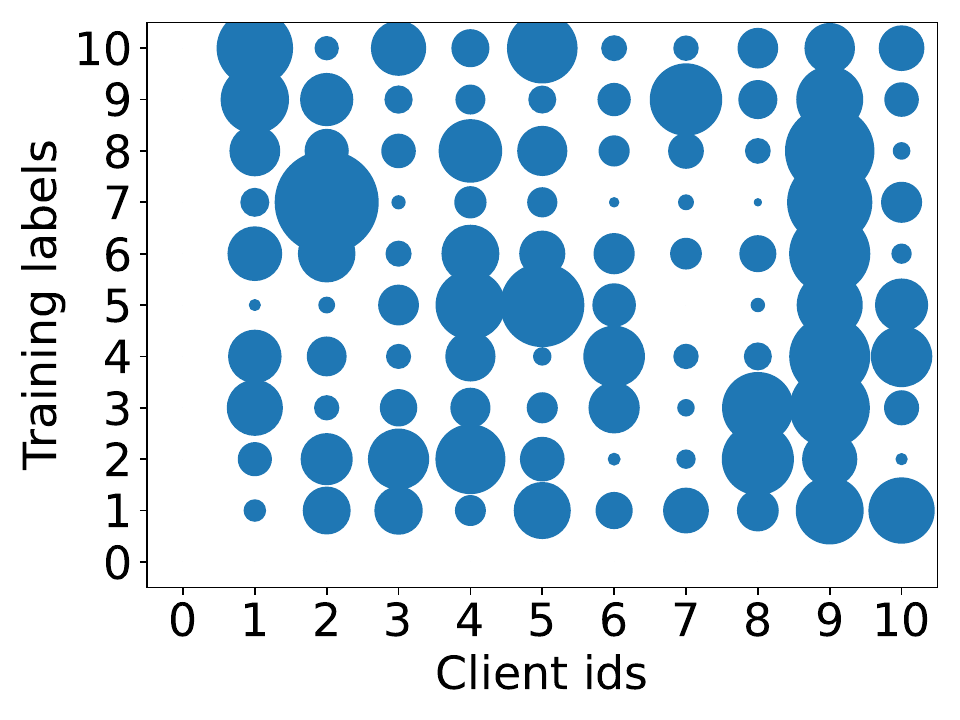}
	}
	\subfigure[CIFAR-10, $\alpha$=0.1, seed=0]{
		\label{fig:Figure_3(d)}
		\includegraphics[scale=0.34]{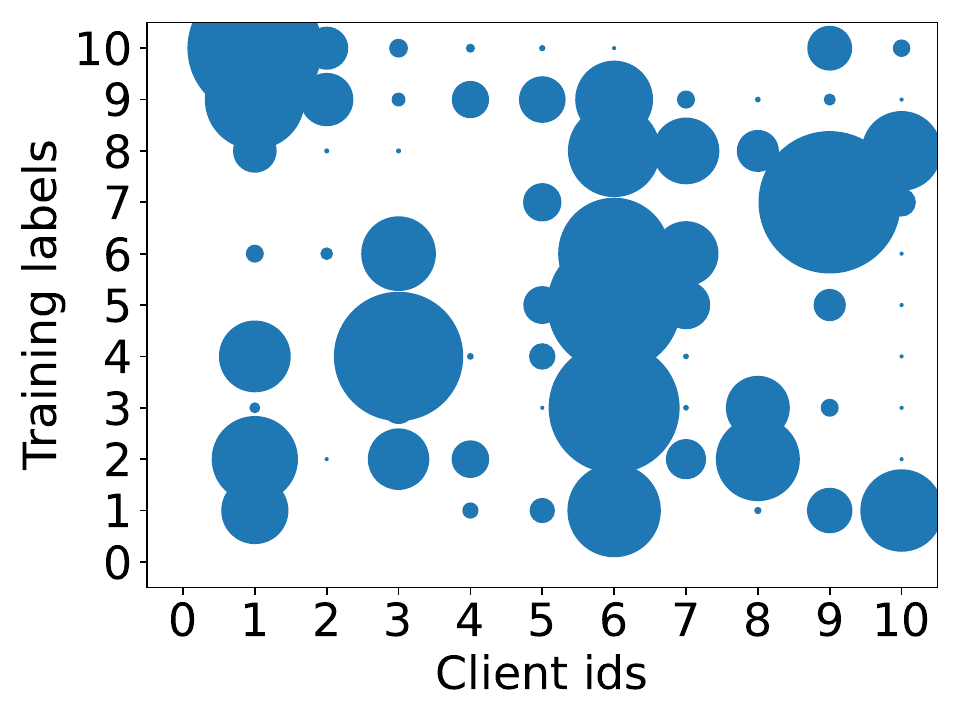}
	}
	\subfigure[CIFAR-10, $\alpha$=0.1, seed=1]{
		\label{fig:Figure_3(e)}
		\includegraphics[scale=0.34]{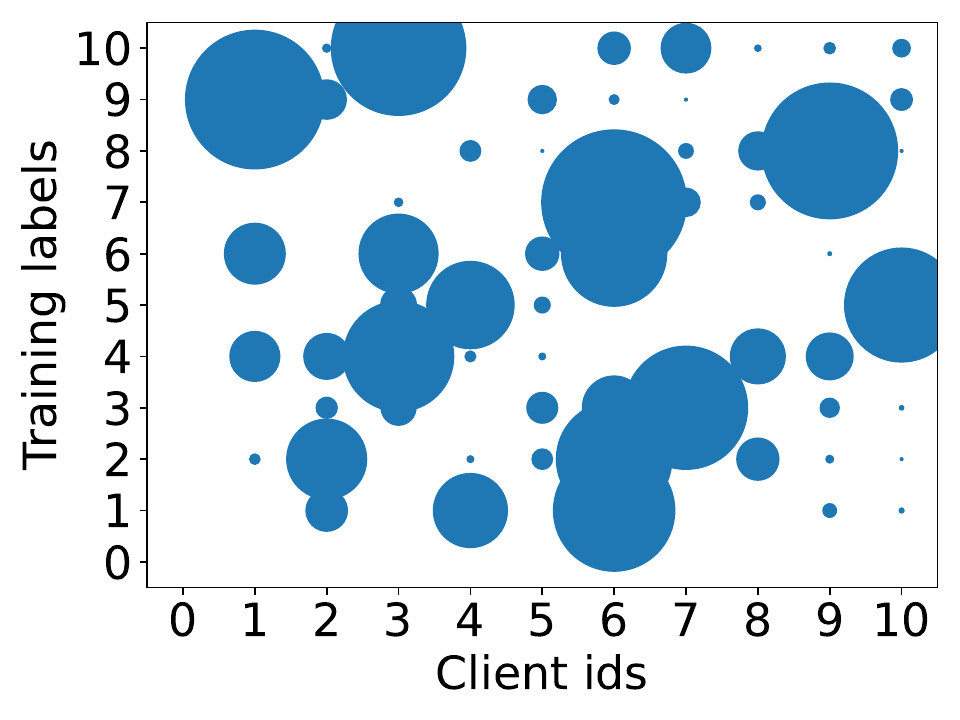}
	}
	\subfigure[CIFAR-10, $\alpha$=0.1, seed=2]{
		\label{fig:Figure_3(f)}
		\includegraphics[scale=0.34]{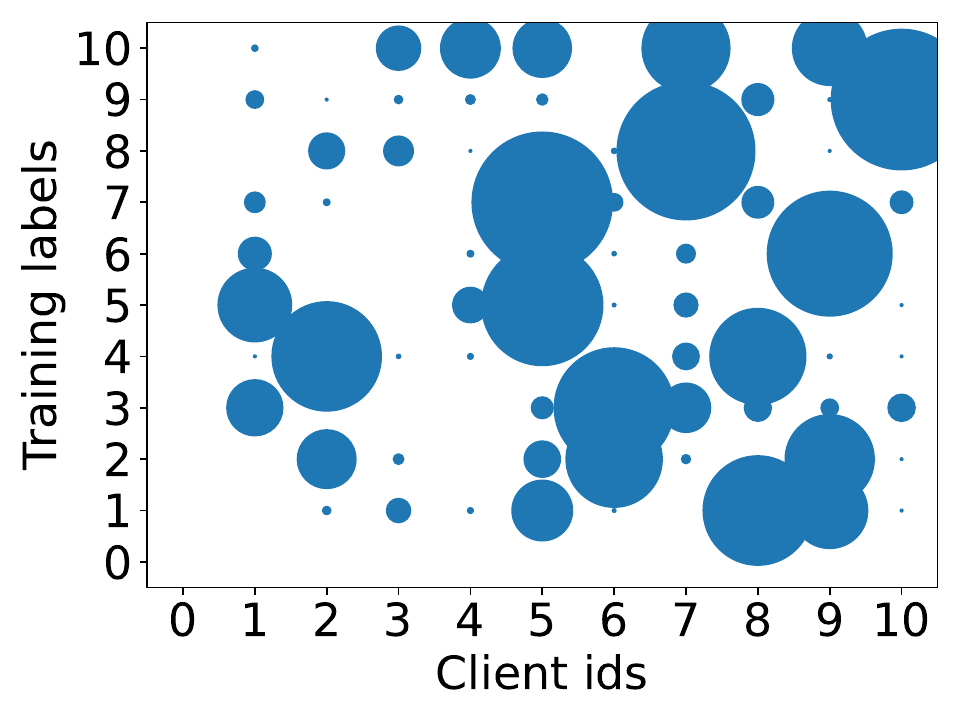}
	}
	\caption{Visualization of data heterogeneity among clients on the CIFAR-10 dataset, where the abscissa denotes the client id, the ordinate denotes the class label, and the size of each scattered point denotes the number of training samples available for each client under each label.}
	\label{fig:Figure_3}
\end{figure*}

\begin{figure*}[!t]
	\centering
	\subfigure[CIFAR-100, $\alpha$=1, seed=0]{
		\label{fig:Figure_4(a)}
		\includegraphics[scale=0.34]{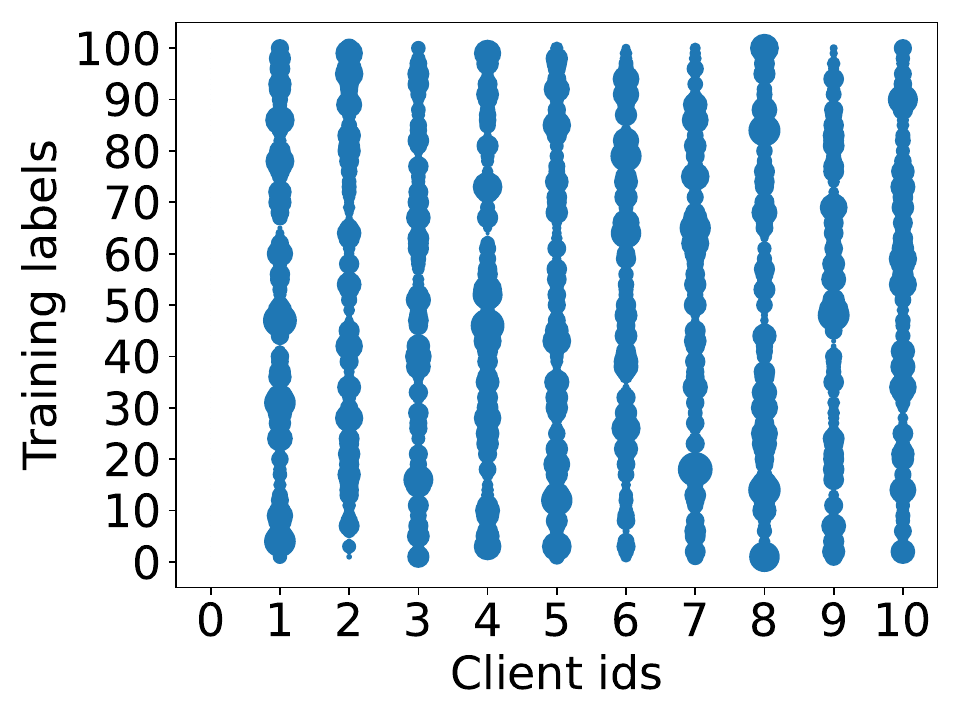}
	}
	\subfigure[CIFAR-100, $\alpha$=1, seed=1]{
		\label{fig:Figure_4(b)}
		\includegraphics[scale=0.34]{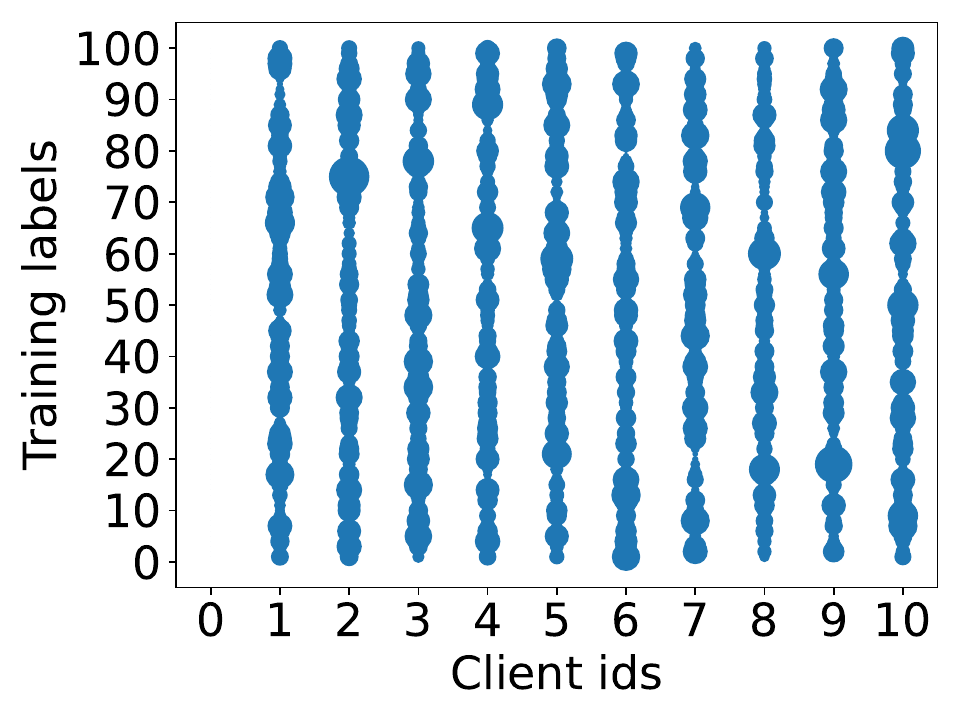}
	}
	\subfigure[CIFAR-100, $\alpha$=1, seed=2]{
		\label{fig:Figure_4(c)}
		\includegraphics[scale=0.34]{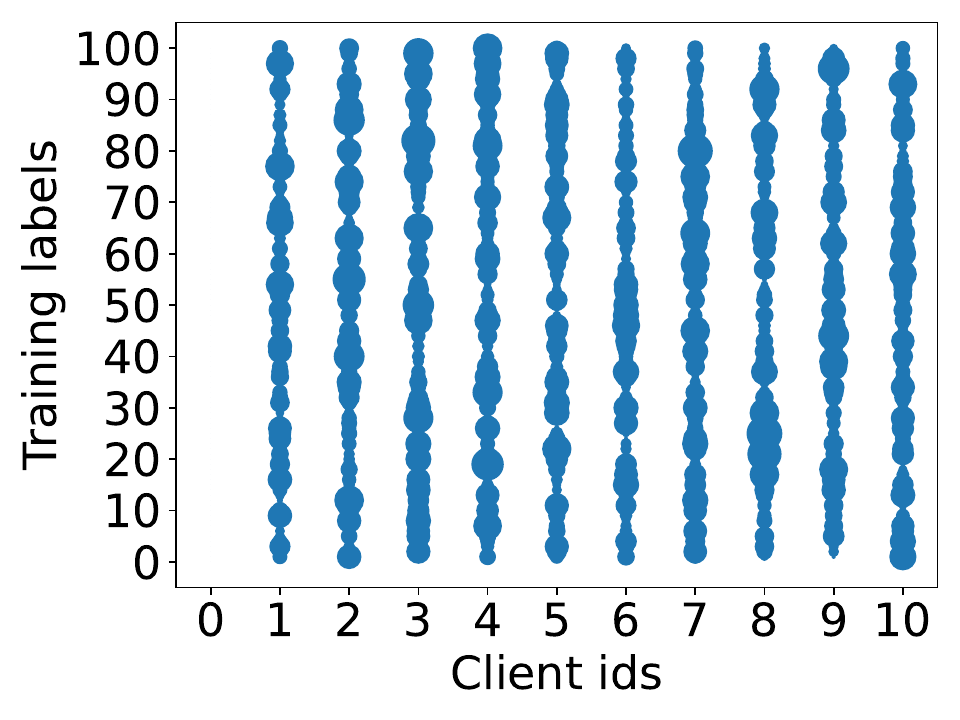}
	}
	\subfigure[CIFAR-100, $\alpha$=0.1, seed=0]{
		\label{fig:Figure_4(d)}
		\includegraphics[scale=0.34]{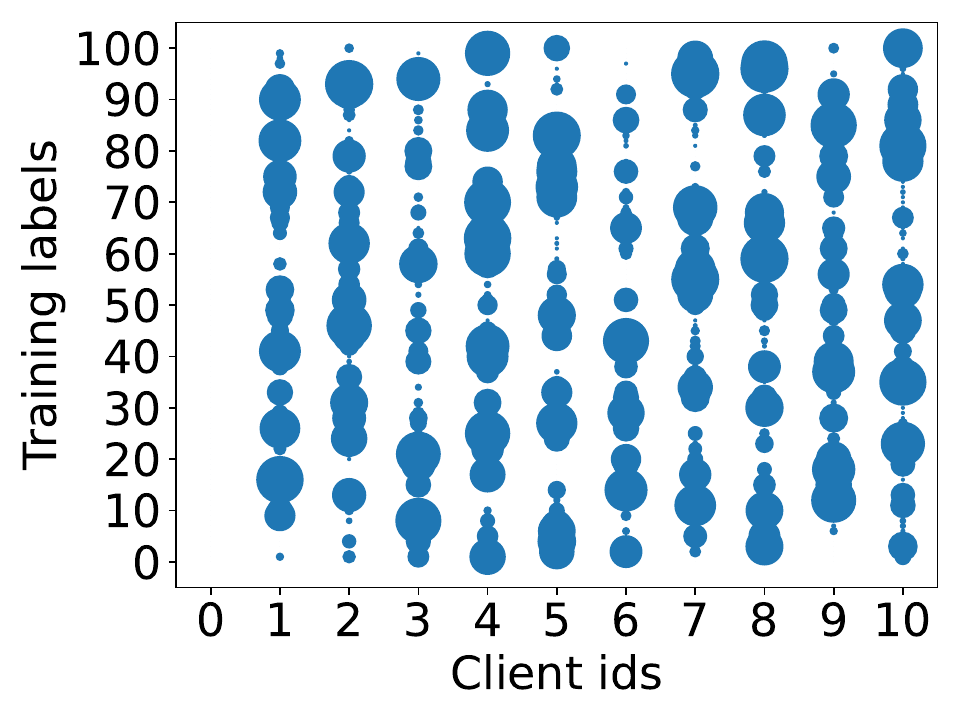}
	}
	\subfigure[CIFAR-100, $\alpha$=0.1, seed=1]{
		\label{fig:Figure_4(e)}
		\includegraphics[scale=0.34]{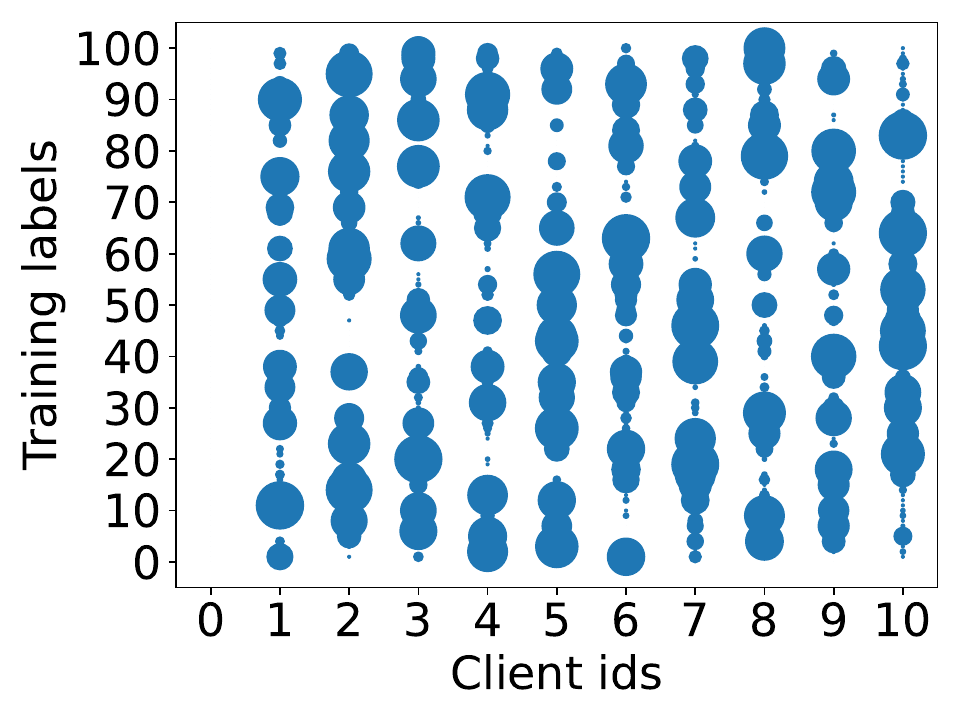}
	}
	\subfigure[CIFAR-100, $\alpha$=0.1, seed=2]{
		\label{fig:Figure_4(f)}
		\includegraphics[scale=0.34]{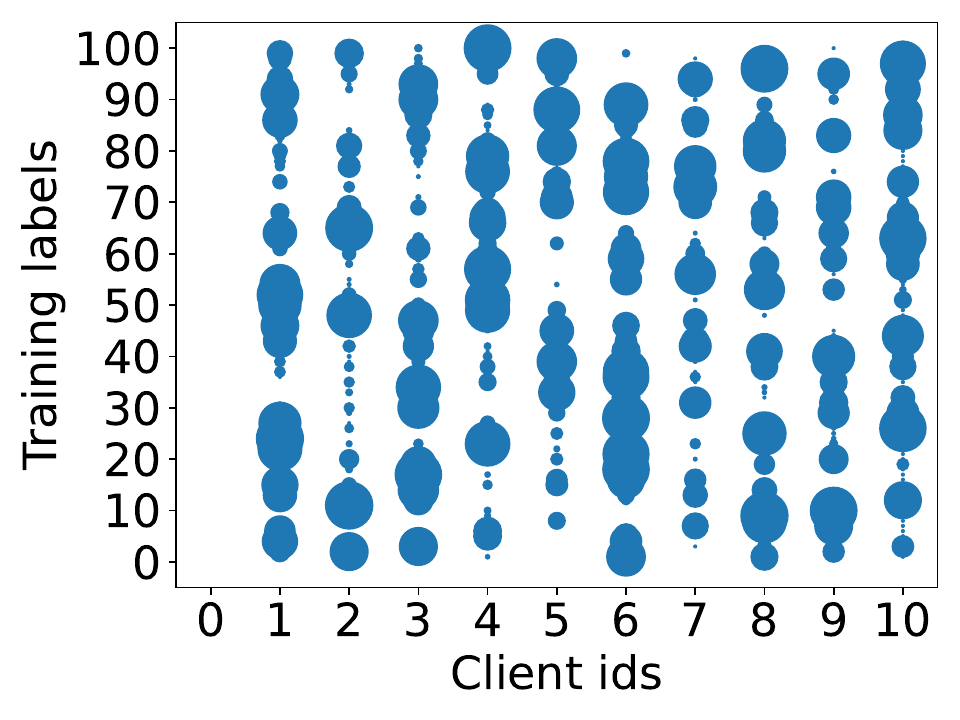}
	}
	\caption{Visualization of data heterogeneity among clients on the CIFAR-100 dataset, where the abscissa denotes the client id, the ordinate denotes the class label, and the size of each scattered point denotes the number of training samples available for each client under each label.}
	\label{fig:Figure_4}
\end{figure*}

\paragraph{Partitioning Details.} To heterogeneously partition the training set of each dataset to $K$ clients as their private data to simulate a realistic data distribution, we use a Dirichlet distribution $Dir(\alpha )$ to sample ${P_c}$ data points for each class $c \in \{ 1,2,..,{\cal C}\} $ from the original training set, and then allocate a proportion of ${P_{c,k}}$ to the client $k \in \{ 1,2,..., K\}$ as its local data belonging to the class $c$. By varying the concentration parameter $\alpha$, we can adjust the degree of non-IID accordingly, where a smaller value of $\alpha$ implies a more heterogeneous data distribution. For intuitive illustration, we provide the visualization of how the training samples of the CIFAR-10 and CIFAR-100 datasets are distributed among $K$=10 clients for different values of $\alpha$ and random seeds in Figures~\ref{fig:Figure_3} and \ref{fig:Figure_4}, respectively.

\subsection{C. Implementation Details}
\begin{table*}[!ht]
	\centering
	\resizebox{\linewidth}{!}{
		\begin{tabular}{cc} 
			\toprule
			\textbf{Layer}                                                                              & \textbf{Output}                         \\ 
			\midrule
			$z \in {\mathbb{R}^{{d_z}}}\sim {\cal N}(0,\bf{1})$                                & $(B,d_z)$                            \\
			$\hat y \in \mathbb{R} \sim {\cal U}(1,\cal C)$,~$\hat y = {\rm{Embedding}}(\hat y)$ & $(B,d_z)$                            \\
			$h{\rm{ = Concat(}}z,\hat y{\rm{)}}$                                                        & $(B,d_z \times 2)$                   \\
			${\rm{Linear(}}h{\rm{)}}$                                                                   & $(B,128 \times (H/4) \times (W/4))$  \\
			${\rm{Reshape(}}h{\rm{)}}$                                                                  & $(B,128,H/4,W/4)$                    \\
			$\rm{BatchNorm2d}(128)$                                                            & $(B,128,H/4,W/4)$                    \\
			$\rm{LeakyReLU(BatchNorm2d}(\rm{ConvTranspose2d(128,128,4,2,1),128}),0.2)$         & $(B,128,H/2,W/2)$                    \\
			$\rm{LeakyReLU(BatchNorm2d}(\rm{ConvTranspose2d(128,64,4,2,1),64}),0.2)$           & $(B,128,H,W)$                        \\
			${\rm{Tanh(BatchNorm2d(Conv2d(64,}}C{\rm{,3,1,1),}}C{\rm{))}}$                           & $(B,C,H,W)$                          \\
			\bottomrule
		\end{tabular}
	}
	\caption{Model architecture of the generator. ${\cal C}$=number of classes, $B$=batch size, $d_z$=dimension of random noises, $H$=height of synthetic images, $W$=width of synthetic images, $C$=channel of synthetic images.}
	\label{tab:Table_7}
\end{table*}
\begin{table*}[!ht]
	\centering
	\resizebox{\linewidth}{!}{
		\begin{tabular}{ccc} 
			\toprule
			\multirow{2.5}{*}{\textbf{Method}} & \multicolumn{2}{c}{\textbf{Formulation}}                                                                                                                                                     \\ 
			\cmidrule{2-3}
			& \textbf{Downlink}                                                                                  & \textbf{Uplink}                                                                                  \\ 
			\midrule
			FedAvg                  & $T \times K \times |\theta |$                                                             & $T \times K \times |\theta _k|$                                                           \\
			\midrule
			MHAT                    & $T \times K \times (|{x_{p}}| + |y_{p}| + |f({x_{p}};\theta )|)$                                & $T \times K \times |f_k({x_{p}};\theta_k )|$                                                \\
			DS-FL                   & $T \times K \times (|{x_{p}}| + |\sum\nolimits_{k \in {\cal K}} {\frac{{{N_k}}}{N}{f_k}({x_{p}};{\theta _k})} |)$                                & $T \times K \times |f_k({x_{p}};\theta _k)|$                                                \\
			\midrule
			FedGen                  & $T \times K \times (|\theta | + |\theta _G|)$                                             & $T \times K \times (|\theta _k| + |c_k|)$                                               \\
			FedFTG                  & $T \times K \times |\theta|$                                                             & $T \times K \times (|\theta _k| + |c_k|)$                                               \\
			DFRD                    & $T \times K \times |\theta|$                                                          & $T \times K \times (|\theta _k| + |c_k|)$                                                \\
			FedZKT                  & $T \times K \times |\theta _k |$                                                             
			& $T \times K \times |\theta _k |$                                                           \\
			\midrule
			FedZGE                  & $T \times K \times (|\hat x| + |\sum\nolimits_{k \in {\cal K}} {\frac{{{N_k}}}{N}{f_k}(\hat x ;{\theta _k})}| + |\hat x + \epsilon{u_i}| \times q)$ & $T \times K \times (|f_k(\hat x;\theta _k)| + |f_k(\hat x + \epsilon{u_i};\theta _k)| \times q)$  \\
			\bottomrule
		\end{tabular}
	}
	\caption{Formulations of communication overhead. $T$=number of communication rounds, $K$=number of clients, $|\theta|$=byte size of the global model, $|{\theta _k}|$=byte size of the $k$-th local model, $|{\theta _G}|$=byte size of the generator, $|f( \cdot ;\theta)|$=byte size of the global model output, $|{f_k}( \cdot ;{\theta _k})|$=byte size of the $k$-th local model output, $|\sum\nolimits_{k \in {\cal K}} {\frac{{{N_k}}}{N}{f_k}( \cdot ;{\theta _k})}|$=byte size of the ensemble output, $|{c_k}|$=byte size of the $k$-th local label statistic, $|{x_{p}}|$=byte size of the auxiliary data, $|{y_{p}}|$=byte size of the corresponding label of auxiliary data, $|\hat x|$=byte size of the synthetic data, $q$=number of random perturbation directions, $|\hat x + \epsilon{u_i}|$=byte size of the synthetic data perturbed along a random direction variable $u_i$.}
	\label{tab:Table_8}
\end{table*}

\paragraph{Hyperparameters.} The experiments are conducted in FL scenarios with data heterogeneity levels of $\alpha$=1 and $\alpha$=0.1, respectively, using $K$=10 clients and $T$=100 communication rounds. In each round, a sampling fraction of $\varepsilon$=1 is used to select $|{{\cal K}}| = \left\lceil {\varepsilon  \cdot K} \right\rceil$ active clients for FL training. Each participating client $k \in \cal K$ trains its on-device model $f_k$ using the Adam optimizer with a learning rate of ${\eta _k}$=0.01 and a batch size of 256 for $E_k$=10 epochs during the local training phase and  $E_k'$=10 epochs during the local distillation phase. The global model $f$ and the generator $G$ are alternately trained for $E=$10 iterations using the Adam optimizer with learning rates of $\eta$=0.01 and ${\eta _G}$=0.001, respectively. Furthermore, the random noise dimension $d_z$ is set to 100, the temperature $\tau$ is set to 5, the scaling factors $\beta_1$, $\beta_2$, and $\beta_3$ are all set to 1, and we sample $q$=10 random perturbation directions with a perturbation size of $\epsilon$=0.001 to reduce variance during gradient estimation. To ensure a fair comparison, we use the CIFAR-100 training set as the auxiliary dataset for experiments on the CIFAR-10 dataset and the CIFAR-10 training set as the auxiliary dataset for experiments on the CIFAR-100 dataset, which results in 50,000 auxiliary data samples for each dataset. Consequently, we set the batch size $B$ to 500, which induces data-free FL approaches to generate 500 synthetic data samples per round, thereby maintaining the same total data size as distillation-based FL approaches. To make data-free FL approaches compatible in both homogeneous and heterogeneous FL settings in consistent with FedZKT, we directly adopt FedFTG and DFRD as data-free methods that transfer knowledge from local models to the global model without parameter aggregation. Particularly, we omit the results of FedAvg and FedGen in the heterogeneous FL setting since they must train the global model through parameter aggregation and thus fail to support heterogeneous local models.

\paragraph{Architecture of Generator.} Following data-free FL approaches, the server-side generator $G$ consists of a label embedding layer, a linear projection layer, a batch normalization layer, two transposed convolutional layers with batch normalization and LeakyReLU activation, and a convolutional layer with batch normalization and Tanh activation. The architecture of the generator is presented in Table~\ref{tab:Table_7}.

\paragraph{Computing Resources.} The experiments are simulated on a single computer equipped with a 13th Gen Intel Core i9-13900K CPU (64G RAM) and an NVIDIA GeForce RTX 4090 GPU (24G RAM), using Python 3.10.10 as the programming environment and PyTorch 2.0.1 as the deep learning platform.

\subsection{D. Formulations of Communication Overhead}
Table~\ref{tab:Table_8} lists the formulations to calculate the communication overhead between the server and clients for different FL approaches, where the downlink and uplink together constitute the communication overhead.

\begin{figure*}[!ht]
	\centering
	\subfigure[CIFAR-10, original]{
		\label{fig:Figure_5(a)}
		\includegraphics[scale=1.165]{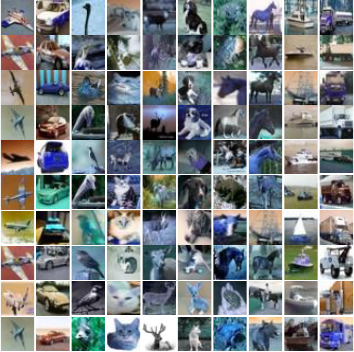}
	}
	\qquad
	\subfigure[CIFAR-10, synthetic]{
		\label{fig:Figure_5(b)}
		\includegraphics[scale=1.165]{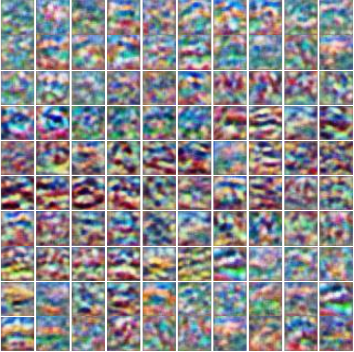}
	}
	\\
	\subfigure[CIFAR-100, original]{
		\label{fig:Figure_5(c)}
		\includegraphics[scale=1.165]{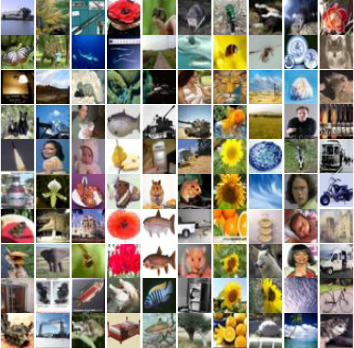}
	}
	\qquad
	\subfigure[CIFAR-100, synthetic]{
		\label{fig:Figure_5(d)}
		\includegraphics[scale=1.165]{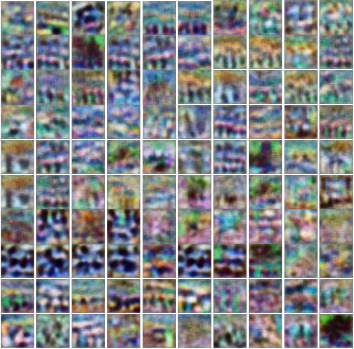}
	}
	\caption{Visualization of original and synthetic data on the CIFAR-10 and CIFAR-100 datasets.}
	\label{fig:Figure_5}
\end{figure*}

\subsection{E. Data Visualization}
Figure~\ref{fig:Figure_5} shows the visualization of the private data and the synthetic data on the CIFAR-10 and CIFAR-100 datasets.

\begin{table*}[!ht]
	\centering
	\resizebox{\linewidth}{!}{
		\begin{tabular}{cccccccccc} 
			\toprule
			\multirow{2.5}{*}{\textbf{\textbf{Method}}} & \multicolumn{3}{c}{$\varepsilon$=0.1}                      & \multicolumn{3}{c}{$\varepsilon$=0.5}                       & \multicolumn{3}{c}{$\varepsilon$=1.0}               \\ 
			\cmidrule{2-10}
			& $\alpha$=1          & $\alpha$=0.1        & Comm           & $\alpha$=1          & $\alpha$=0.1        & Comm            & $\alpha$=1          & $\alpha$=0.1        & Comm    \\ 
			\cmidrule{2-10}
			FedAvg                                    & 45.12$\pm$0.15          & 33.30$\pm$0.43          & 41.65          & 62.58$\pm$0.27          & 51.87$\pm$0.42          & 208.27          & 75.14$\pm$0.40          & 64.20$\pm$0.30          & 416.55  \\ 
			\midrule
			MHAT                                      & 41.29$\pm$0.23          & 30.98$\pm$0.41          & 2.88           & 59.34$\pm$0.40          & 47.65$\pm$0.46          & 14.40           & 72.14$\pm$0.17          & 61.13$\pm$0.60          & 28.81   \\
			DS-FL                                     & 40.87$\pm$0.10          & 30.63$\pm$0.58          & 2.88           & 58.79$\pm$0.31          & 47.13$\pm$0.21          & 14.40           & 71.88$\pm$0.30          & 60.69$\pm$0.41          & 28.80   \\ 
			\midrule
			FedGen                                    & 45.56$\pm$0.34          & 34.58$\pm$0.39          & 45.46          & 64.76$\pm$0.28          & 52.34$\pm$0.57          & 227.30          & 75.45$\pm$0.10          & 65.18$\pm$0.15          & 454.61  \\
			FedFTG                                    & 45.72$\pm$0.21          & 34.92$\pm$0.46          & 41.65          & 64.98$\pm$0.11          & 52.51$\pm$0.40          & 208.27          & 75.88$\pm$0.18          & 65.44$\pm$0.44          & 416.55  \\
			DFRD                                      & 45.90$\pm$0.13          & 35.40$\pm$0.40          & 41.65          & 65.05$\pm$0.13          & 52.75$\pm$0.37          & 208.27          & \textbf{76.31$\pm$0.29} & \textbf{65.76$\pm$0.29} & \textbf{416.55}  \\
			FedZKT                                    & 45.26$\pm$0.11          & 34.89$\pm$0.32          & 41.65          & 64.86$\pm$0.35          & 51.94$\pm$0.29          & 208.27          & 75.60$\pm$0.13          & 64.85$\pm$0.50          & 416.55  \\ 
			\midrule
			FedZGE                                    & \textbf{46.34$\pm$0.19} & \textbf{35.84$\pm$0.54} & \textbf{31.58} & \textbf{65.13$\pm$0.42} & \textbf{52.84$\pm$0.44} & \textbf{157.92} & 76.12$\pm$0.24          & 65.51$\pm$0.32          & 315.83  \\
			\bottomrule
		\end{tabular}
	}
	\caption{Test accuracy (\%) and communication overhead (GB) on the CIFAR-10 dataset under different client populations using ResNet-18.}
	\label{tab:Table_9}
\end{table*}
\begin{table*}[!t]
	\centering
	\resizebox{\linewidth}{!}{
		\begin{tabular}{cccccccccc} 
			\toprule
			\multirow{2.5}{*}{\textbf{\textbf{Method}}} & \multicolumn{3}{c}{$\varepsilon$=0.1}                      & \multicolumn{3}{c}{$\varepsilon$=0.5}                       & \multicolumn{3}{c}{$\varepsilon$=1.0}                        \\ 
			\cmidrule{2-10}
			& $\alpha$=1          & $\alpha$=0.1        & Comm           & $\alpha$=1          & $\alpha$=0.1        & Comm            & $\alpha$=1          & $\alpha$=0.1        & Comm             \\ 
			\midrule
			FedAvg                   & 36.24$\pm$0.24          & 24.68$\pm$0.45          & 41.83          & 46.12$\pm$0.09          & 30.66$\pm$0.38          & 209.13          & 51.78$\pm$0.34          & 37.40$\pm$0.58          & 418.27           \\ 
			\midrule
			MHAT                     & 33.10$\pm$0.16          & 20.74$\pm$0.33          & 3.05           & 41.67$\pm$0.13          & 26.89$\pm$0.52          & 15.24           & 47.31$\pm$0.14          & 33.11$\pm$0.47          & 30.48            \\
			DS-FL                    & 32.87$\pm$0.11          & 20.33$\pm$0.19          & 3.05           & 41.22$\pm$0.25          & 26.45$\pm$0.59          & 15.24           & 47.04$\pm$0.29          & 32.80$\pm$0.27          & 30.47            \\ 
			\midrule
			FedGen                   & 36.45$\pm$0.34          & 24.81$\pm$0.56          & 45.65          & 46.34$\pm$0.37          & 30.54$\pm$0.26          & 228.25          & 52.08$\pm$0.10          & 37.56$\pm$0.54          & 456.50           \\
			FedFTG                   & 36.61$\pm$0.17          & 24.99$\pm$0.18          & 41.83          & 46.62$\pm$0.11          & 31.10$\pm$0.44          & 209.13          & 52.24$\pm$0.19          & 37.72$\pm$0.32          & 418.27           \\
			DFRD                     & 36.83$\pm$0.26          & 25.22$\pm$0.37          & 41.83          & 46.71$\pm$0.24          & 31.27$\pm$0.49          & 209.13          & \textbf{52.41$\pm$0.18} & \textbf{37.95$\pm$0.43} & \textbf{418.27}  \\
			FedZKT                   & 36.34$\pm$0.21          & 24.89$\pm$0.44          & 41.83          & 46.36$\pm$0.20          & 30.89$\pm$0.36          & 209.13          & 52.18$\pm$0.28          & 37.51$\pm$0.12          & 418.27           \\ 
			\midrule
			FedZGE                   & \textbf{37.25$\pm$0.12} & \textbf{25.67$\pm$0.26} & \textbf{32.59} & \textbf{46.84$\pm$0.28} & \textbf{31.36$\pm$0.37} & \textbf{162.94} & 52.32$\pm$0.17          & 37.88$\pm$0.32          & 325.89           \\
			\bottomrule
		\end{tabular}
	}
	\caption{Test accuracy (\%) and communication overhead (GB) on the CIFAR-100 dataset under different client populations using ResNet-18.}
	\label{tab:Table_10}
\end{table*}

\subsection{F. Extended Experimental Results}
\paragraph{Scalability to Larger Client Populations.} To investigate the scalability of different FL approaches in FL scenarios with larger client populations, we conduct additional experiments on $K$=50 clients with different sampling fractions $\varepsilon$=\{0.1, 0.5, 1.0\}, which results in FL systems with 5, 25, and 50 participating clients, respectively, and then evaluate the test accuracy and communication overhead of different FL approaches using ResNet-18 on the CIFAR-10 and CIFAR-100 datasets in Tables~\ref{tab:Table_9} and \ref{tab:Table_10}, respectively. It can be observed that the performance of all FL approaches manifests a descending trend since distributing the entire training set to more clients reduces the amount of local data available for each client. However, the performance of all FL approaches progressively improves as the value of $\varepsilon$ increases since the more the number of clients participated in the FL training process, the more local data becomes available. At the same time, more participating clients incurs a linear increase in communication overhead for all FL approaches. FedZGE makes a trade-off between distillation-based FL and data-free FL by introducing a communication overhead that is higher than that of distillation-based FL approaches but significantly lower than that of data-free FL approaches in exchange for performance that is much higher than that of distillation-based FL approaches and comparable to that of data-free FL approaches. It is worth noting that the superiority of FedZGE diminishes as the number of clients increases since the cumulative error in gradient estimation is amplified as the number of local models to be estimated. Nevertheless, FedZGE can still achieve comparable performance with less communication overhead in FL scenarios with larger client populations.

\begin{figure}[!t]
	\centering
	\subfigure[CIFAR-10, $\alpha$=1]{
		\label{fig:Figure_6(a)}
		\includegraphics[scale=0.255]{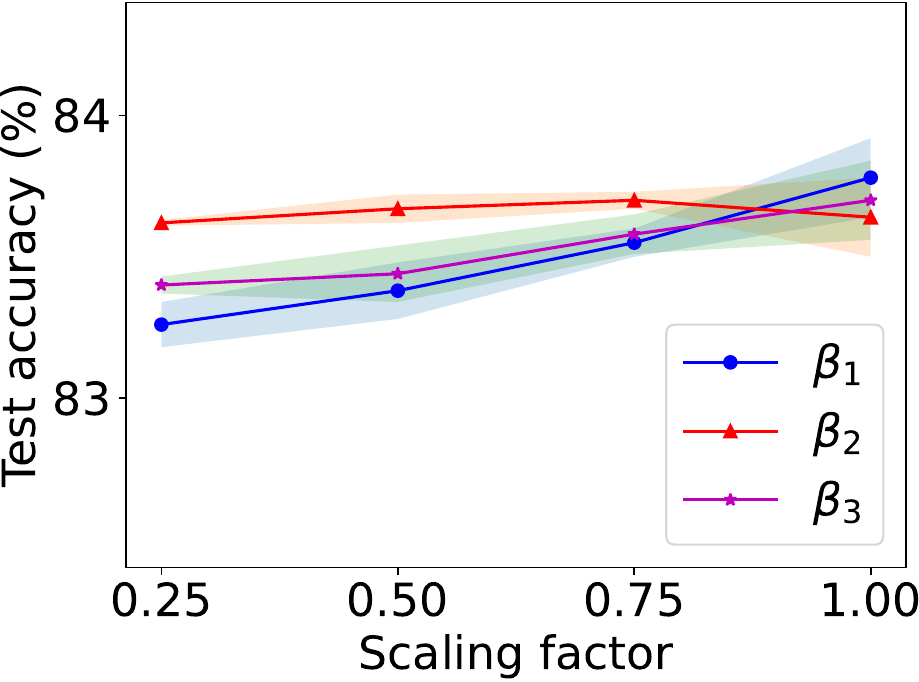}
	}
	\subfigure[CIFAR-10, $\alpha$=0.1]{
		\label{fig:Figure_6(b)}
		\includegraphics[scale=0.255]{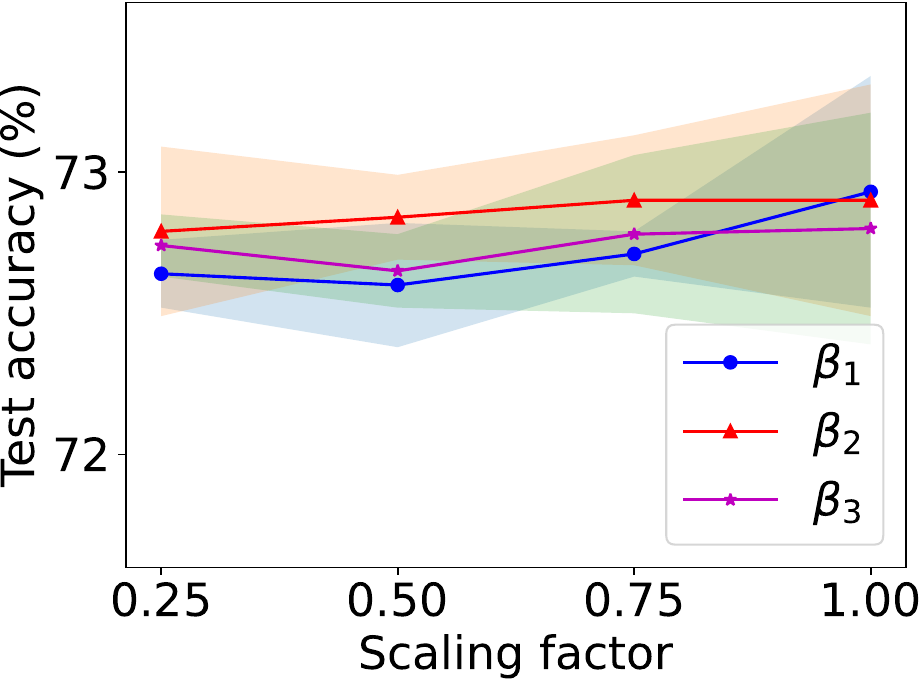}
	}
	\\
	\subfigure[CIFAR-100, $\alpha$=1]{
		\label{fig:Figure_6(c)}
		\includegraphics[scale=0.255]{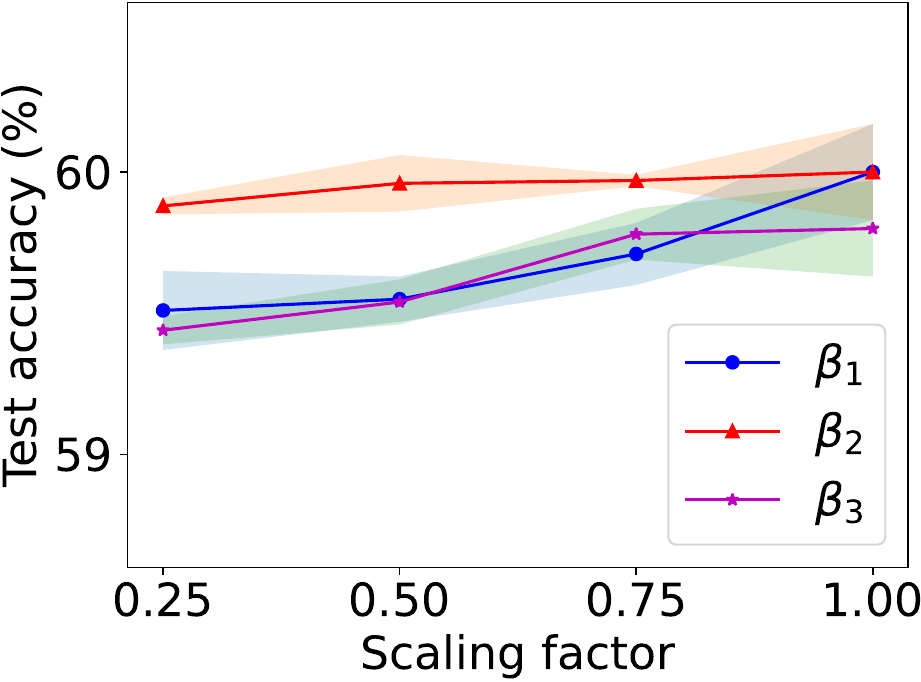}
	}
	\subfigure[CIFAR-100, $\alpha$=0.1]{
		\label{fig:Figure_6(d)}
		\includegraphics[scale=0.255]{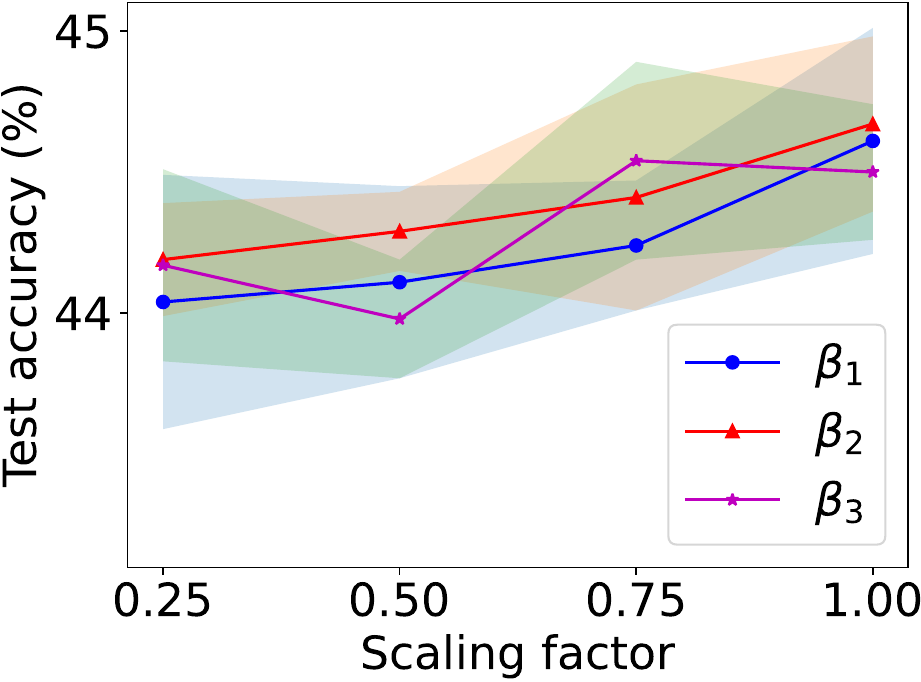}
	}
	\caption{Ablation study of FedZGE with different scaling factors on the CIFAR-10 and CIFAR-100 datasets using ResNet-50.}
	\label{fig:Figure_6}
\end{figure}

\paragraph{Effects of Scaling Factors.} To explore the effects of the three scaling factors $\beta_1$, $\beta_2$, and $\beta_3$ introduced in Eq.~\eqref{eq:Equation_10}, which control the contribution of adversarial loss, diversity loss, and information entropy loss to the training of the generator, respectively, we select each scaling factor from a set of common values \{0.25, 0.50, 0.75, 1.00\} and then evaluate the resulting test accuracy using ResNet-50 on the CIFAR-10 and CIFAR-100 datasets in Figure~\ref{fig:Figure_6}. It can be observed that the values of these three scaling factors are generally positively correlated with their contributions for a given range of choices, \emph{i.e.}, the larger the value, the better the performance. Among them, the adversarial loss controlled by $\beta_1$ has a more significant impact on performance, while the diversity loss controlled by $\beta_2$ has a minor impact on performance. Moreover, the information entropy loss controlled by $\beta_3$ exhibits a more significant effect on the CIFAR-100 dataset, suggesting that it helps control the balance of the generated synthetic data samples across different classes. Overall, FedZGE maintains a relatively consistent performance across different scaling factors.

\end{document}